\documentclass{article}

\usepackage{nips10submit_e,times}
\usepackage{graphicx}
\usepackage{subfig}
\usepackage{amssymb,amsmath,amsthm}
\usepackage{bm}
\usepackage[round]{natbib}

\newtheorem{theorem}{Theorem}
\newtheorem{corollary}[theorem]{Corollary}
\newtheorem{lemma}[theorem]{Lemma}
\newtheorem{proposition}[theorem]{Proposition}

\newtheorem{definition}[theorem]{Definition}

\newcommand{\N}{\mathbb{N}}   
\newcommand{\R}{\mathbb{R}}   
\newcommand{\U}{\mathcal{U}}  
\newcommand{\V}{\mathcal{V}}  
\DeclareMathOperator{\Exp}{\mathbb{E}}  
\DeclareMathOperator{\Vol}{\mathbf{Vol}}  
\DeclareMathOperator{\rank}{rank}  
\newcommand{\ud}{\mathrm{d}}  
\newcommand{\x}{\mathbf{x}}   
\renewcommand{\u}{\mathbf{u}}   
\renewcommand{\b}{\mathbf{b}}   
\newcommand{\e}{\mathbf{e}}   
\newcommand{\y}{\mathbf{y}}   
\newcommand{\X}{\mathbf{X}}   
\newcommand{\Y}{\mathbf{Y}}   
\renewcommand{\S}{\mathbf{S}}   
\newcommand{\A}{\mathbf{A}}   
\newcommand{\W}{\mathbf{W}}   
\newcommand{\Z}{\mathbf{Z}}   
\newcommand{\z}{\mathbf{z}}   
\newcommand{\F}{\mathbf{F}}   
\newcommand{\ie}{\emph{i.e.}}   

\title{Estimation of R\'enyi Entropy and Mutual Information Based on Generalized Nearest-Neighbor Graphs}

\author{
D\'avid P\'al \\
Department of Computing Science \\
University of Alberta \\
Edmonton, AB, Canada \\
\texttt{dpal@cs.ualberta.ca} \\
\And
Barnab\'as P\'oczos%
\\
School of Computer Science \\
Carnegie Mellon University \\
Pittsburgh, PA, USA \\
\texttt{poczos@ualberta.ca}
\And
Csaba Szepesv\'ari \\
Department of Computing Science \\
University of Alberta \\
Edmonton, AB, Canada \\
\texttt{szepesva@ualberta.ca}
}

\nipsfinalcopy

\begin{document}
\maketitle

\renewcommand{\thefootnote}{\fnsymbol{footnote}}
\renewcommand{\thefootnote}{\arabic{footnote}}

\begin{abstract}
We present simple and computationally efficient nonparametric estimators of
R\'enyi entropy and mutual information based on an i.i.d. sample drawn from an
unknown, absolutely continuous distribution over $\R^d$.  The estimators are
calculated as the sum of $p$-th powers of the Euclidean lengths of the edges of
the `generalized nearest-neighbor' graph of the sample and the empirical copula
of the sample respectively. For the first time,
we prove the almost sure consistency of these estimators and
 upper bounds on their rates of convergence, the latter of which under
the assumption that the density underlying the sample is Lipschitz continuous.
Experiments demonstrate their usefulness in independent subspace analysis.
\end{abstract}

\section{Introduction}

We consider the nonparametric problem of estimating R\'enyi $\alpha$-entropy and mutual
information (MI) based on a finite sample drawn from an unknown, absolutely
continuous distribution over $\R^d$.  
There are many applications that make use of such estimators, of which we list
a few to give the reader a taste: Entropy estimators can be used for
goodness-of-fit testing \citep{vasicek76test,goria05new}, parameter estimation
in semi-parametric models \citep{Wolsztynski85minimum}, studying fractal
random walks \citep{Alemany94fractal}, and texture classification
\citep{hero2002aes,hero02alpha}.  Mutual information estimators have been used
in feature selection \citep{peng05feature}, clustering
\citep{aghagolzadeh07hierarchical}, causality detection
\citep{Hlavackova07causality}, optimal experimental design
\citep{lewi07realtime,poczos09identification},
fMRI data processing \citep{chai09exploring}, prediction of protein structures
\citep{adami04information}, or boosting and facial expression recognition
\citep{Shan05conditionalmutual}.  Both entropy estimators and mutual
information estimators have been used for independent component and subspace
analysis
\citep{radical03,poczos05geodesic,Hulle08constrained,szabo07undercomplete_TCC},
and image registration  \citep{kybic06incremental,hero2002aes,hero02alpha}. For
further applications, see
\citet{Leonenko-Pronzato-Savani2008,WKV2009Survey}.

In a na\"ive approach to R\'enyi entropy and mutual information estimation, one
could use the so called ``plug-in'' estimates. These are based on the obvious
idea that since entropy and mutual information are determined solely by the
density $f$ (and its marginals), it suffices to first estimate the density using one's
favorite density estimate which is then ``plugged-in'' into the formulas
defining entropy and mutual information. The density is, however, a nuisance
parameter which we do \emph{not} want to estimate.  Density estimators have
tunable parameters and we may need cross validation to achieve good
performance.

The entropy estimation algorithm considered here is \emph{direct}---it does not
build on density estimators. It is based on $k$-nearest-neighbor (NN) graphs
with a fixed $k$.  A variant of these estimators, where each sample point is
connected to its $k$-th nearest neighbor only, were recently studied by
\citet{goria05new} for Shannon entropy estimation (\ie{} the special case
$\alpha=1$) and \citet{Leonenko-Pronzato-Savani2008} for R\'enyi $\alpha$-entropy
estimation. They proved the \emph{weak} consistency of their estimators
under certain conditions.
However, their proofs contain some errors, and it is not obvious how to fix them.  Namely,
\citet{Leonenko-Pronzato-Savani2008} apply the generalized Helly-Bray theorem,
while  \citet{goria05new} apply the inverse Fatou lemma under conditions when these
theorems do not hold. This latter error originates from the article of
\cite{kozachenko87statistical}, and this mistake can also be found in
\citet{Wang-Kulkarni-Verdu2009}.

The first main contribution of this paper is to give a
correct proof of consistency of these estimators. Employing a very different
proof techniques than the papers mentioned above, we show that these estimators
are, in fact, \emph{strongly} consistent provided that the unknown density $f$
has bounded support and $\alpha \in (0,1)$. At the same time, we allow for more
general nearest-neighbor graphs, wherein as opposed to connecting each point
only to its $k$-th nearest neighbor, we allow each point to be connected to an
arbitrary subset of its $k$ nearest neighbors. Besides adding generality, our
numerical experiments seem to suggest that connecting each sample point to all
its $k$ nearest neighbors improves the rate of convergence of the estimator.

The second major contribution of our paper is that we prove a finite-sample
high-probability bound on the error (\ie{} the rate of convergence) of our
estimator provided that $f$ is Lipschitz. According to the best of our knowledge, this
is the very first result that gives a rate for the estimation of R\'enyi
entropy. The closest to our result in this respect is the work by
\citet{tsybakov96rootn} who proved the root-$n$ consistency of an estimator of
the {\em Shannon} entropy and only in {\em one} dimension.

The third contribution is a \emph{strongly} consistent estimator of
R\'enyi mutual information that is based on NN graphs and the empirical copula
transformation~\citep{dedecker07weak}. This result is proved for $d\ge
3$~\footnote{Our result for R\'enyi entropy estimation holds for $d=1$ and
$d=2$, too.} and $\alpha \in (1/2,1)$. This
builds upon and extends the previous work of
\citet{Poczos-Kirshner-Szepesvari2010} where instead of NN graphs, the minimum
spanning tree (MST) and the shortest tour through the sample (\ie{} the
traveling salesman problem, TSP) were used, but it was only conjectured that NN
graphs can be applied as well.

There are several advantages of using $k$-NN graph over MST and TSP (besides
the obvious conceptual simplicity of $k$-NN): On a serial computer the $k$-NN
graph can be computed somewhat faster than MST and much faster than the TSP
tour.  Furthermore, in contrast to MST and TSP, computation of $k$-NN can be
easily parallelized.  Secondly, for different values of $\alpha$, MST and TSP
need to be recomputed since the distance between two points is the $p$-th power
of their Euclidean distance where $p=d(1-\alpha)$. However, the $k$-NN graph
does not change for different values of $p$, since $p$-th power is a monotone
transformation, and hence the estimates for multiple values of $\alpha$ can be
calculated without the extra penalty incurred by the recomputation of the
graph. This can be advantageous \emph{e.g.} in intrinsic dimension estimators of
manifolds \citep{costa03entropic}, where $p$ is a free parameter, and thus one
can calculate the estimates efficiently for a few different parameter values.

The fourth major contribution is a proof of a finite-sample high-probability
error bound (\ie{} the rate of convergence) for our mutual information estimator
which holds under the assumption that the copula of $f$ is Lipschitz. According
to the best of our knowledge, this is the first result that gives a rate for
the estimation of R\'enyi mutual information.

The toolkit for proving our results derives from the deep literature of
Euclidean functionals, see, \citep{Steele1997,Yukich1998}.  In particular, our
strong consistency result uses a theorem due to  \citet{Redmond-Yukich1996}
that essentially states that any quasi-additive power-weighted Euclidean
functional can be used as a strongly consistent estimator of R\'enyi entropy
(see also \citealt{HeMi99}). We also make use of a result due to
\citet{Koo-Lee2007}, who proved a rate of convergence result that holds under
more stringent conditions.
Thus, the main thrust of the present work is showing that  these conditions hold for $p$-power
weighted nearest-neighbor graphs.
Curiously enough,
up to now, no one has shown this, except for the case when $p=1$, which is
studied in Section~8.3 of \citep{Yukich1998}. However, the condition $p=1$
gives results only for $\alpha=1-1/d$.

All proofs and supporting lemmas can be found in the appendix.  In the main
body of the paper, we focus on clear explanation of R\'enyi entropy and mutual
information estimation problems, the estimation algorithms and the statements
of our converge results.

Additionally, we report on two numerical experiments. In the first experiment,
we compare the empirical rates of convergence of our estimators with our
theoretical results and plug-in estimates.  Empirically, the NN methods are the
clear winner.  The second experiment is an illustrative application of mutual
information estimation to an Independent Subspace Analysis (ISA) task.

The paper is organized as follows: In the next section, we formally define
R\'enyi entropy and R\'enyi mutual information and the problem of their
estimation.  Section~\ref{section:nearest-neighbors} explains the `generalized
nearest neighbor' graphs. This graph is then used in
Section~\ref{section:entropy} to define our R\'enyi entropy estimator. In the
same section, we state a theorem containing our convergence results  for this
estimator (strong consistency and rates).  In
Section~\ref{section:information}, we explain the copula transformation, which
connects R\'enyi entropy with R\'enyi mutual information. The copula
transformation together with the R\'enyi entropy estimator from
Section~\ref{section:entropy} is used to build an estimator of R\'enyi mutual
information. We conclude this section with a theorem stating the convergence
properties of the estimator (strong consistency and rates).
Section~\ref{section:experiments} contains the numerical experiments.  We
conclude the paper by a detailed discussion of further related work in
Section~\ref{section:discussion}, and a list of open problems and directions
for future research in Section~\ref{section:conclusions}.

\section{The Formal Definition of the Problem}
\label{section:formal-problem}

R\'enyi entropy and R\'enyi mutual information of $d$ real-valued random
variables\footnote{We use superscript for indexing dimension coordinates.}
$\X=(X^1, X^2, \dots, X^d)$  with joint density $f:\R^d \to \R$ and marginal
densities $f_i:\R \to \R$, $1 \le i \le d$, are defined for any real parameter
$\alpha$ assuming the underlying integrals exist. For $\alpha \neq 1$, R\'enyi
entropy and R\'enyi mutual information are defined respectively as\footnote{The
base of the logarithms in the definition is not important; any base strictly
bigger than $1$ is allowed.  Similarly as with Shannon entropy and mutual
information, one traditionally uses either base $2$ or $e$. In this paper, for
definitiveness, we stick to base $e$.}
\begin{align}
\label{equation:renyi-entropy}
H_\alpha(\X) & = H_\alpha(f) = \frac{1}{1-\alpha} \log \int_{\R^d} f^\alpha(x^1, x^2, \dots, x^d) \ \ud (x^1, x^2, \dots, x^d) \;  ,
\\
\label{equation:renyi-mutual-information}\hspace*{-1cm}
I_\alpha(\X)  & = I_\alpha(f) = \frac{1}{\alpha-1} \log \int_{\R^d} f^\alpha(x^1, x^2, \dots, x^d) \left( \prod_{i=1}^d f_i(x^i) \right)^{\hspace*{-1mm} 1-\alpha} \hspace*{-2mm} \ud (x^1, x^2, \dots, x^d) .
\end{align}
For $\alpha = 1$ they are defined by the limits $H_1 = \lim_{\alpha \to
1} H_\alpha$ and $I_1 = \lim_{\alpha \to 1} I_\alpha$. In fact,
Shannon (differential) entropy and the Shannon mutual information
are just special cases of R\'enyi entropy and R\'enyi mutual information
with $\alpha = 1$.

The goal of this paper is to present estimators of R\'enyi entropy
\eqref{equation:renyi-entropy} and R\'enyi information
\eqref{equation:renyi-mutual-information} and study their convergence
properties. To be more explicit, we consider the problem where we are given
i.i.d. random variables $\X_{1:n}=(\X_1, \X_2, \dots, \X_n)$ where each $\X_j =
(X_{j}^1, X_{j}^2, \dots, X_{j}^d)$ has density $f:\R^d \to \R$ and marginal
densities $f_i:\R \to \R$ and our task is to construct an estimate $\widehat
H_\alpha(\X_{1:n})$ of $H_\alpha(f)$ and an estimate $\widehat
I_\alpha(\X_{1:n})$ of $I_\alpha(f)$ using the sample $\X_{1:n}$.

\section{Generalized Nearest-Neighbor Graphs}
\label{section:nearest-neighbors}

The basic tool to define our estimators is the generalized nearest-neighbor graph
and more specifically the sum of the $p$-th powers of Euclidean lengths of its
edges.

Formally, let $V$ be a finite set of points in an Euclidean space $\R^d$ and
let $S$ be a finite non-empty set of positive integers; we denote by $k$ the
maximum element of $S$. We define the \emph{generalized nearest-neighbor graph}
$NN_S(V)$ as a directed graph on $V$.  The edge set of $NN_S(V)$ contains for
each $i \in S$ an edge from each vertex $\x \in V$ to its $i$-th nearest
neighbor.  That is, if we sort $V \setminus \{\x\} = \{\y_1, \y_2, \dots,
\y_{|V|-1}\}$ according to the Euclidean distance to $\x$ (breaking ties
arbitrarily): $\|\x - \y_1\| \le \|\x - \y_2\| \le \dots \le \|\x-\y_{|V|-1}\|$
then $\y_i$ is the $i$-th nearest-neighbor of $\x$ and for each $i \in S$ there
is an edge from $\x$ to $\y_i$ in the graph.

For $p \ge 0$ let us denote by $L_p(V)$ the sum of the $p$-th powers of Euclidean
lengths of its edges.  Formally,
\begin{equation}
\label{equation:nn-functional}
L_p(V) = \sum_{(\x,\y) \in E(NN_S(V))} \|\x-\y\|^p \; ,
\end{equation}
where $E(NN_S(V))$ denotes the edge set of $NN_S(V)$. We intentionally hide
the dependence on $S$ in the notation $L_p(V)$. For the rest of the paper, the
reader should think of $S$ as a fixed but otherwise arbitrary finite non-empty
set of integers, say, $S=\{1,3,4\}$.

The following is a basic result about $L_p$. The proof can be found in the appendix.
\begin{theorem}[Constant $\gamma$]
\label{theorem:constant-gamma}
Let $\X_{1:n} = (\X_1, \X_2, \dots, \X_n)$ be an i.i.d. sample from the uniform distribution
over the $d$-dimensional unit cube $[0,1]^d$. For any $p \ge 0$ and any finite non-empty set $S$ of positive integers
there exists a constant $\gamma > 0$ such that
\begin{equation}
\label{equation:constant-gamma}
\lim_{n \to \infty} \frac{L_p(\X_{1:n})}{n^{1-p/d}} = \gamma \qquad {a.s.}
\end{equation}
\end{theorem}

The value of $\gamma$ depends on $d,p,S$ and, except for special cases, an
analytical formula for its value is not known. This causes a minor problem
since the constant $\gamma$ appears in our estimators. A simple and effective
way to deal with this problem is to generate a large i.i.d. sample $\X_{1:n}$
from the uniform distribution over $[0,1]^d$ and estimate $\gamma$ by the
empirical value of $L_p(\X_{1:n})/n^{1-p/d}$.

\section{An Estimator of R\'enyi Entropy} \label{section:entropy}

We are now ready to present an estimator of R\'enyi entropy based on the
generalized nearest-neighbor graph.  Suppose we are given an i.i.d. sample
$\X_{1:n} = (\X_1, \X_2, \dots, \X_n)$ from a distribution $\mu$ over $\R^d$
with density $f$. We estimate entropy $H_\alpha(f)$ for $\alpha \in (0,1)$ by
\begin{equation}\label{equation:entropy-estimator}
\widehat H_\alpha(\X_{1:n}) =
\frac{1}{1-\alpha} \log \frac{L_p(\X_{1:n})}{\gamma n^{1-p/d}} \qquad \text{where} \quad p=d(1-\alpha),
\end{equation}
and $L_p(\cdot)$ is the sum of $p$-th powers of Euclidean
lengths of edges of the nearest-neighbor graph $NN_S(\cdot)$ for some finite non-empty $S
\subset \N^+$ as defined by equation~\eqref{equation:nn-functional}. The constant
$\gamma$ is the same as in Theorem~\ref{theorem:constant-gamma}.

The following theorem is our main result about the estimator $\widehat H_\alpha$.
It states that $\widehat H_\alpha$ is strongly consistent and gives upper bounds
on the rate of convergence. The proof of theorem is in the appendix.

\begin{theorem}[Consistency and Rate for $\widehat H_\alpha$]
\label{thm:mainentropy}
Let $\alpha \in (0,1)$. Let $\mu$ be an absolutely continuous distribution over $\R^d$ with bounded support
and let $f$ be its density. If $\X_{1:n} = (\X_1, \X_2, \dots, \X_n)$ is an i.i.d. sample from $\mu$ then
\begin{eqnarray}
\lim_{n \to \infty} \widehat H_\alpha(\X_{1:n}) = H_\alpha(f) \qquad \text{a.s.} \label{equation:entropy-as}
\end{eqnarray}
Moreover, if $f$ is Lipschitz then for any $\delta > 0$ with probability at least $1-\delta$,
\begin{eqnarray}
\label{equation:entropy-rate}
\left| \widehat H_\alpha(\X_{1:n}) - H_\alpha(f) \right|
\le
\begin{cases}
O\left( n^{-\frac{d-p}{d(2d-p)}}(\log(1/\delta))^{1/2-p/(2d)} \right), & \text{if $0 < p < d-1$} \; ; \\
O\left( n^{-\frac{d-p}{d(d+1)}}(\log(1/\delta))^{1/2-p/(2d)} \right), & \text{if $d-1 \le p < d$} \; .
\end{cases}
\end{eqnarray}
\end{theorem}

\section{Copulas and Estimator of Mutual Information}
\label{section:information}

Estimating mutual information is slightly more complicated than estimating entropy.
We start with a basic property of mutual information which we call \emph{rescaling}. It states
that if $h_1, h_2, \dots, h_d:\R \to \R$ are arbitrary strictly increasing functions, then
\begin{equation}
I_\alpha(h_1(X^1), h_2(X^2), \dots, h_d(X^d)) = I_\alpha(X^1, X^2, \dots, X^d) \; .
\end{equation}
A particularly clever choice is $h_j = F_j$ for all $1 \le j \le d$,
where $F_j$ is the cumulative
distribution function (c.d.f.) of $X^j$. With this choice, the marginal
distribution of $h_j(X^j)$ is the uniform distribution over $[0,1]$ assuming
that $F_j$, the c.d.f. of $X^j$, is continuous.  Looking at the
definition of $H_\alpha$ and $I_\alpha$ we see that
\begin{equation*}
I_\alpha(X^1, X^2, \dots, X^d)
= I_\alpha(F_1(X^1), F_2(X^2), \dots, F_d(X^d))
= -H_\alpha(F_1(X^1), F_2(X^2), \dots, F_d(X^d)) \; .
\end{equation*}
In other words, calculation of mutual information can be reduced to the
calculation of entropy provided that marginal c.d.f.'s $F_1, F_2, \dots, F_d$
are known. The problem is, of course, that these are not known and need to be
estimated from the sample. We will use empirical c.d.f.'s
$(\widehat{F}_1,\widehat{F}_2,\dots,\widehat{F}_d)$ as their estimates. Given
an i.i.d. sample $\X_{1:n}=(\X_1,\X_2,\dots,\X_n)$ from distribution $\mu$ and
with density $f$, the empirical c.d.f's are defined as
$$
\widehat F_j(x) = \frac{1}{n} | \{ i  : 1 \le i \le n, \ x \le X_{i}^j  \} | \qquad \text{for } x \in \R, \ 1 \le j \le d \; .
$$
Introduce the compact notation
$\F:\R^d \to [0,1]^d$,
$\widehat \F:\R^d \to [0,1]^d$,
\begin{eqnarray}
\label{equation:copula-transformation}
 \F(x^1, x^2, \dots, x^d) &=& (F_1(x^1),  F_2(x^2), \dots,  F_d(x^d)) \qquad \text{for } (x^1, x^2, \dots, x^d) \in \R^d\;;\\
\widehat \F(x^1, x^2, \dots, x^d) &=& (\widehat F_1(x^1), \widehat F_2(x^2), \dots, \widehat F_d(x^d)) \qquad \text{for } (x^1, x^2, \dots, x^d) \in \R^d\;.
\label{equation:empirical-copula-transformation}
\end{eqnarray}

Let us call the maps $\F$, $\widehat \F$ the \emph{copula transformation}, and
the \emph{empirical copula transformation}, respectively.  The joint
distribution of $\F(\X) = (F_1(X^1),F_2(X^2),\ldots,F_d(X^d))$ is called the copula of
$\mu$, and the sample $(\widehat\Z_1,\widehat\Z_2,\ldots,\widehat\Z_n)=
(\widehat\F(\X_1),\widehat\F(\X_2),\ldots,\widehat\F(\X_n))$ is called the
empirical copula \citep{dedecker07weak}. Note that $j$-th coordinate  of $\widehat\Z_i$ equals
$$\widehat Z_i^j=\frac{1}{n} \rank(X_{i}^j, \{ X_{1}^j, X_{2}^j, \dots, X_{n}^j
\}) \; ,$$ where $\rank(x,A)$ is the number of element of $A$ less than or
equal to $x$. Also, observe that the random variables
$\widehat\Z_1,\widehat\Z_2,\ldots,\widehat\Z_n$ are not even independent!
Nonetheless, the empirical copula $(\widehat \Z_1,\widehat \Z_2,\ldots,\widehat
\Z_n)$ is a good approximation of an i.i.d. sample $(\Z_1,\Z_2,\ldots,\Z_n)=
(\F(\X_1), \F(\X_2), \dots, \F(\X_n))$ from the copula of $\mu$.
Hence, we estimate the R\'enyi mutual information ${I}_{\alpha}$ by
\begin{eqnarray}
\widehat{I}_{\alpha}(\X_{1:n}) = - \widehat H_{\alpha}(\widehat \Z_1, \widehat \Z_2, \dots, \widehat \Z_n),
\end{eqnarray}
where $\widehat H_{\alpha}$ is defined by \eqref{equation:entropy-estimator}.
The following theorem is our main result about the estimator $\widehat I_\alpha$.
It states that $\widehat I_\alpha$ is strongly consistent and gives upper bounds
on the rate of convergence. The proof of this theorem can be found in the appendix.

\begin{theorem}[Consistency and Rate for $\widehat I_\alpha$]
\label{thm:maininformation}
Let $d\geq 3$ and $\alpha=1-p/d \in (1/2,1)$. Let $\mu$ be an absolutely continuous distribution over $\R^d$ with
density $f$. If $\X_{1:n} = (\X_1, \X_2, \dots, \X_n)$ is an i.i.d. sample from $\mu$ then
$$
\lim_{n \to \infty} \widehat I_\alpha(\X_{1:n})= I_\alpha(f) \qquad \text{a.s.}
$$
Moreover, if the density of the copula of $\mu$ is Lipschitz, then for any $\delta > 0$ with probability at least $1-\delta$,
\begin{eqnarray*}
\left|  \widehat{I}_{\alpha}(\X_{1:n}) \ - I_{\alpha}(f) \right|
&\le&\begin{cases}
O\left( \max\{n^{-\frac{d-p}{d(2d-p)}},n^{-p/2+p/d} \}
(\log(1/\delta))^{1/2} \right), & \text{if $0 < p \le 1$}\;; \\
O\left( \max\{n^{-\frac{d-p}{d(2d-p)}},n^{-1/2+p/d}\}(\log(1/\delta))^{1/2} \right),  & \text{if $1 \le p \le d-1$}\;; \\
O\left( \max\{n^{-\frac{d-p}{d(d+1)}},n^{-1/2+p/d}\}(\log(1/\delta))^{1/2} \right),  & \text{if $d-1 \le p <d$}\;.
\end{cases}
\end{eqnarray*}
\end{theorem}

\section{Experiments}
\label{section:experiments}

In this section we show two numerical experiments to support our theoretical
results about the convergence rates, and to demonstrate the applicability of
the proposed R\'enyi mutual information estimator, $\widehat{I}_\alpha$.

\subsection{The Rate of Convergence}

\begin{figure}[h]
\begin{center}
\subfloat[3D uniform]{\includegraphics[width=0.33\linewidth]{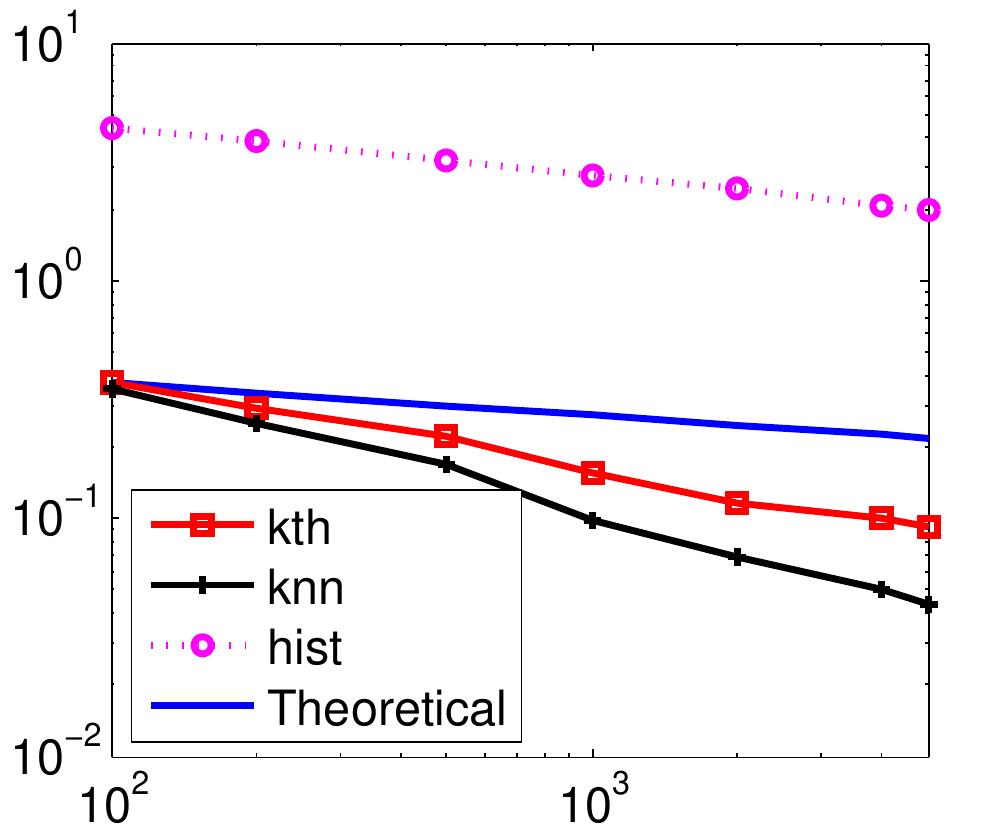}\label{fig:3Duniform}}
\subfloat[3D Gaussian]{\includegraphics[width=0.33\linewidth]{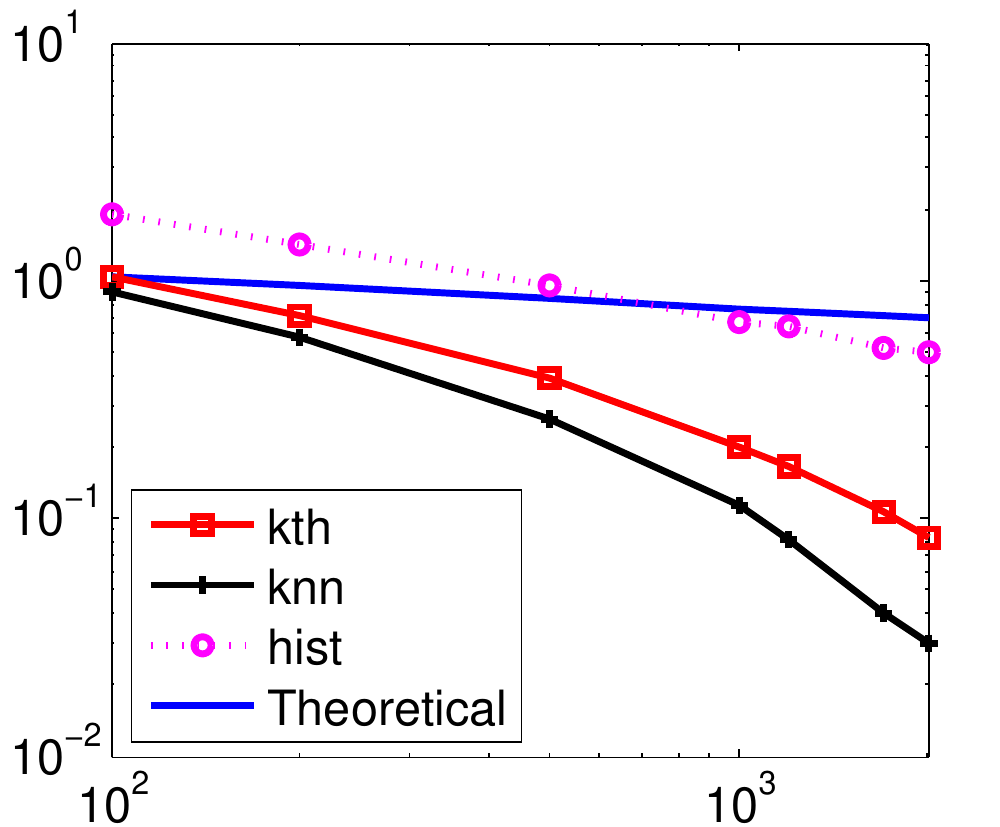}\label{fig:3DGauss}}
\subfloat[20D Gaussian]{\includegraphics[width=0.33\linewidth]{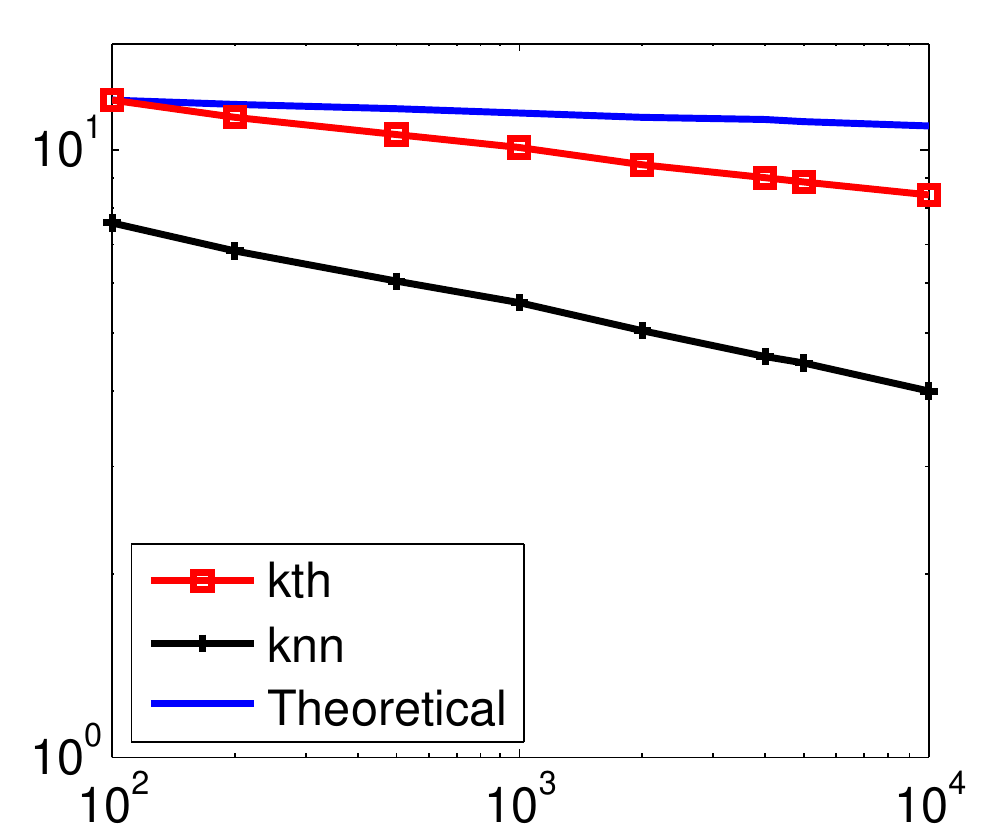}\label{fig:20DGauss}}
\caption{Error of the estimated R\'enyi informations in the number of samples.}
\label{figure:rates}
\end{center}
\end{figure}

In our first experiment (Fig.~\ref{figure:rates}), we demonstrate that the
derived rate is indeed an upper bound on the convergence rate.
Figure~\ref{fig:3Duniform}-\ref{fig:20DGauss} show the estimation error of
$\widehat{I}_{\alpha}$ as a function of the sample size. Here, the underlying
distribution was a 3D uniform, a 3D Gaussian, and a 20D Gaussian with randomly
chosen nontrivial covariance matrices, respectively. In these experiments
$\alpha$ was set to $0.7$. For the estimation we used $S=\{3\}$ (kth) and
$S=\{1,2,3\}$ (knn) sets.  Our results also indicate that these estimators
achieve better performances than the histogram based plug-in estimators (hist).
The number and the sizes of the bins were determined with the rule of
\cite{scott79optimal}. The histogram based estimator is not shown in the 20D
case, as in this large dimension it is not applicable in practice. The figures
are based on averaging 25 independent runs, and they also show the theoretical
upper bound (Theoretical) on the rate derived in
Theorem~\ref{thm:maininformation}.  It can be seen that the theoretical rates
are rather conservative. We think that this is because the theory allows for
quite irregular densities, while the densities considered in this experiment
are very nice.

\subsection{Application to Independent Subspace Analysis}
\label{section:ISA}

\begin{figure}[h]
\begin{center}
\subfloat[Original]{\includegraphics[width=0.29\linewidth]{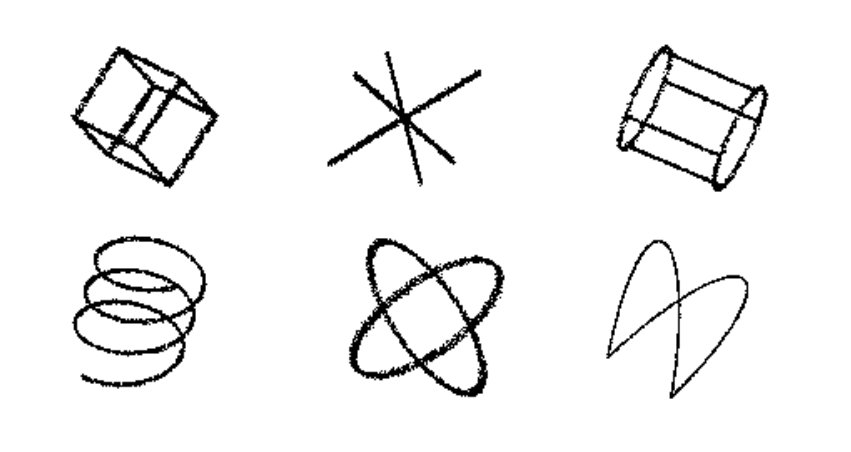}\label{fig:original-ISA}}
\subfloat[Mixed]{\includegraphics[width=0.29\linewidth]{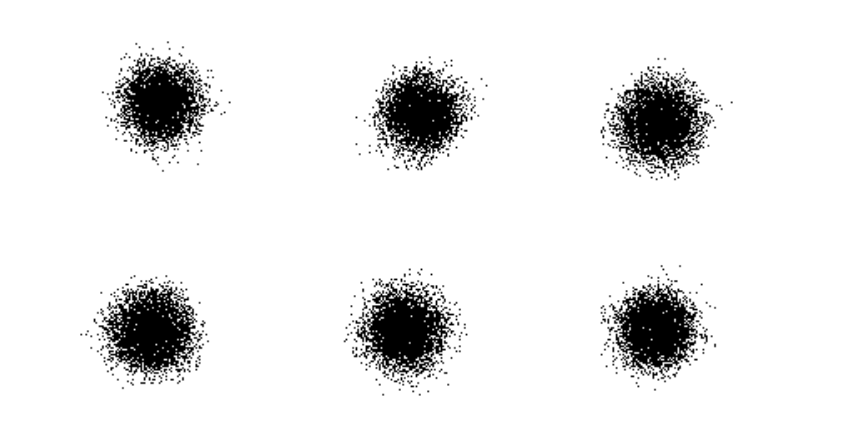}\label{fig:mixed-ISA}}
\subfloat[Estimated]{\includegraphics[width=0.28\linewidth]{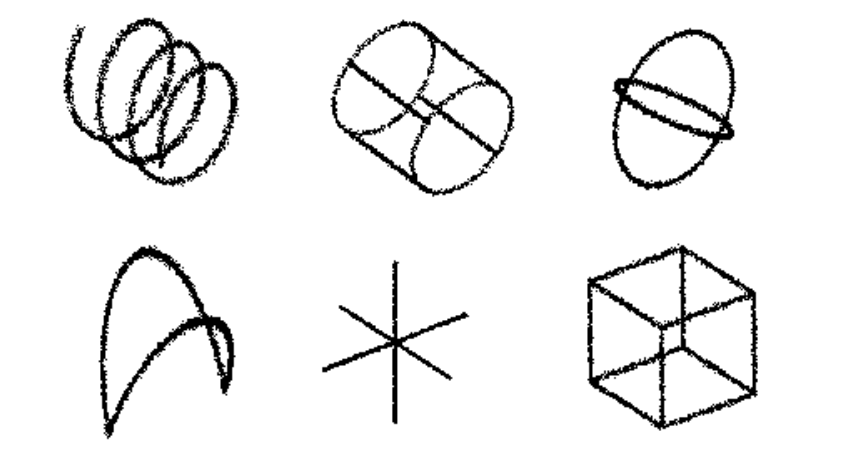}\label{fig:estimated-ISA}}
\subfloat[Hinton]{\includegraphics[width=0.13\linewidth]{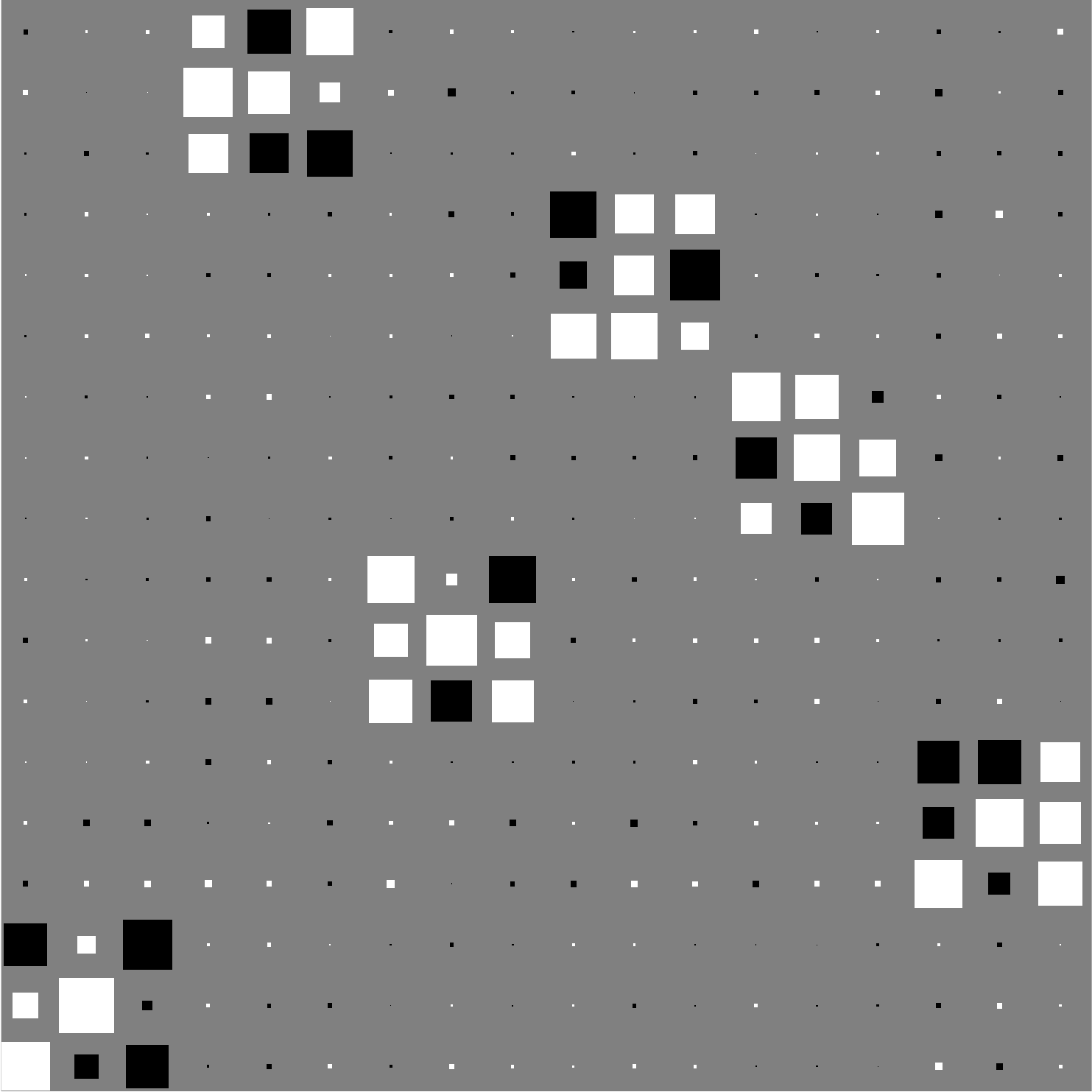}\label{fig:hinton-ISA}}
\caption{ISA experiment for six $3$-dimensional sources.}
\label{figure:ISA}
\end{center}
\end{figure}

An important application of dependence estimators is the Independent Subspace
Analysis problem \citep{cardoso98multidimensional}. This problem is a
generalization of the Independent Component Analysis (ICA), where we assume the
independent sources are multidimensional vector valued random variables. The
formal description of the problem is as follows. We have
$\S=(\S^1;\ldots;\S^m)\in\mathbb{R}^{dm}$, $m$ independent $d$-dimensional
sources, \ie~$\S^i \in \mathbb{R}^d$, and
$I(\S^1,\ldots,\S^m)=0$.\footnote{Here we need the generalization of MI to
multidimensional quantities, but that is obvious by simply replacing the 1D
marginals by $d$-dimensional ones.} In the ISA statistical model we assume that
$\S$ is hidden, and only $n$ i.i.d. samples from $\X=\A\S$ are available for
observation, where $\A\in \mathbb{R}^{q \times dm}$ is an unknown invertible
matrix with full rank and $q \ge dm$. Based on $n$ i.i.d. observation of $\X$,
our task is to estimate the hidden sources $\S^i$ and the mixing matrix $\A$.
Let the estimation of $\S$ be denoted by
$\Y=(\Y^1;\ldots;\Y^m)\in\mathbb{R}^{dm}$, where $\Y=\W\X$.  The goal of ISA is
to calculate $\textrm{argmin}_{\W} I(\Y^1,\ldots,\Y^m)$, where $\W\in
\mathbb{R}^{dm\times q}$ is a matrix with full rank.  Following the ideas of
\citet{cardoso98multidimensional}, this ISA problem can be solved by first
preprocessing the observed quantities $\X$ by a traditional ICA algorithm which
provides us $\W_{ICA}$ estimated separation matrix\footnote{for simplicity we
used the FastICA algorithm in our experiments \citep{ICAbook01}}, and then
simply grouping the estimated ICA components into ISA subspaces by maximizing
the sum of the MI in the estimated subspaces, that is we have to find a
permutation matrix $\mathbf{P} \in \{0,1\}^{dm\times dm}$ which solves
\begin{align}
\max_{\mathbf{P}}\sum_{j=1}^m I(Y_1^j,Y_2^j,\dots,Y_{d}^j) \label{e:J_ISA} \; .
\end{align}
where $\Y=\mathbf{P}\W_{ICA}\X$.  We used the proposed copula based information
estimation, $\widehat{I}_\alpha$ with $\alpha=0.99$ to approximate the Shannon
mutual information, and we chose $S=\{1,2,3\}$. Our experiment shows that this
ISA algorithm using the proposed MI estimator can indeed provide good
estimation of the ISA subspaces. We used a standard ISA benchmark dataset from
\cite{szabo07undercomplete_TCC}; we generated 2,000 i.i.d. sample points on 3D
geometric wireframe distributions from 6 different sources independently from
each other. These sampled points can be seen in Fig.~\ref{fig:original-ISA},
and they represent the sources, $\S$. Then we mixed these sources by a randomly
chosen invertible matrix $\A \in \mathbb{R}^{18 \times 18}$. The six
3-dimensional projections of $\X=\A\S$ observed quantities are shown in
Fig.~\ref{fig:mixed-ISA}. Our task was to estimate the original sources $\S$
using the sample of the observed quantity $\X$ only. By estimating the MI in
\eqref{e:J_ISA}, we could recover the original subspaces as it can be seen in
Fig.~\ref{fig:estimated-ISA}. The successful subspace separation is shown in
the form of Hinton diagrams as well, which is the product of the estimated ISA
separation matrix $\W=\mathbf{P}\W_{ICA}$ and $\A$. It is a block permutation
matrix if and only if the subspace separation is perfect
(Fig.~\ref{fig:hinton-ISA}).

\section{Further Related Works}
\label{section:discussion}

As it was pointed out earlier, in this paper we heavily built on the results
known from the theory of Euclidean functionals
\citep{Steele1997,Redmond-Yukich1996,Koo-Lee2007}.  However, now we can be more
precise about earlier work concerning nearest-neighbor based Euclidean
functionals: The closest to our work is Section~8.3 of \citet{Yukich1998},
where the case of $NN_S$ graph based $p$-power weighted Euclidean functionals
with $S=\{1,2,\ldots,k\}$ and $p=1$ was investigated.

Nearest-neighbor graphs have first been proposed for Shannon entropy estimation
by \citet{kozachenko87statistical}. In particular, in the mentioned work only
the case of $NN_S$ graphs with $S=\{1\}$ was considered.  More recently,
\citet{goria05new} generalized this approach to $S=\{k\}$ and proved the
resulting estimator's weak consistency under some conditions on the density.
The estimator in this paper has a form quite similar to that of ours:
$$
\tilde{H}_1=\log(n-1)-\psi(k)+\log\left(\frac{2\pi^{d/2}}{d\Gamma(d/2)}\right)+\frac{d}{n} \, \sum_{i=1}^n \log\|\e_i\| \; .
$$
Here $\psi$ stands for the digamma function, and $\e_i$ is the directed edge
pointing from $\X_i$ to its $k^{th}$ nearest-neighbor.  Comparing this
with~\eqref{equation:entropy-estimator}, unsurprisingly, we find that the main
difference is the use of the logarithm function instead of $|\cdot|^p$ and the
different normalization.
As mentioned before, \citet{Leonenko-Pronzato-Savani2008} proposed an estimator that uses the $NN_S$ graph with $S=\{k\}$ for the purpose of estimating the R\'enyi entropy. Their estimator takes the form
$$
\tilde{H}_{\alpha}=\frac{1}{1-\alpha} \log\left(\frac{n-1}{n}V_d^{1-\alpha}C_k^{1-\alpha}
\sum_{i=1}^n\frac{\|\e_i\|^{d(1-\alpha)}}{(n-1)^{\alpha}}
\right) \; ,
$$
where $\Gamma$ stands for the Gamma function,
$C_k=\left[\frac{\Gamma(k)}{\Gamma(k+1-\alpha)}\right]^{1/(1-\alpha)}$ and
$V_d=\pi^{d/2}\Gamma(d/2+1)$ is the volume of the $d$-dimensional unit ball,
and  again $\e_i$ is the directed edge in the $NN_S$ graph starting from node
$\X_i$ and pointing to the $k$-th nearest node. Comparing this estimator
with~\eqref{equation:entropy-estimator}, it is apparent that it is
(essentially) a special case of our $NN_S$ based estimator.  From the results
of \citet{Leonenko-Pronzato-Savani2008} it is obvious that the constant
$\gamma$ in~\eqref{equation:entropy-estimator} can be found in analytical form
when $S=\{k\}$. However, we kindly warn the reader again that the proofs of
these last three cited articles
\citep{kozachenko87statistical,goria05new,Leonenko-Pronzato-Savani2008} contain
a few errors, just like the \cite{Wang-Kulkarni-Verdu2009} paper for
\emph{KL divergence} estimation from two samples. \citet{Kraskov04estimating}
also proposed a $k$-nearest-neighbors based estimator for the {\em Shannon}
mutual information estimation, but the theoretical properties of their
estimator are unknown.

\section{Conclusions and Open Problems}
\label{section:conclusions}

We have studied R\'enyi entropy and mutual information estimators
based on $NN_S$ graphs. The estimators were shown to be strongly consistent.
In addition, we derived upper bounds on their convergence rate under some
technical conditions. Several open problems remain unanswered:

An important open problem is to understand how the choice of the set $S \subset
\N^+$ affects our estimators. Perhaps, there exists a way to choose $S$ as a
function of the sample size $n$ (and $d,p$) which strikes the optimal balance
between the bias and the variance of our estimators.

Our method can be used for estimation of Shannon entropy and mutual information
by simply using $\alpha$ close to $1$. The open problem is to come up with a
way of choosing $\alpha$, approaching $1$, as a function of the sample size $n$
(and $d,p$) such that the resulting estimator is consistent and converges as
rapidly as possible. An alternative is to use the logarithm function in place
of the power function.  However, the theory would need to be changed significantly
to show that the resulting estimator remains
strongly consistent.

In the proof of consistency of our mutual information estimator
$\widehat I_\alpha$
we used Kiefer-Dvoretzky-Wolfowitz theorem to handle the effect of the
inaccuracy of the empirical copula transformation. Our particular use of the
theorem seems to restrict $\alpha$ to the interval $(1/2,1)$ and the dimension
to values larger than $2$. Is there a better way to estimate the error caused by the
empirical copula transformation and prove consistency of the estimator for
a larger range of $\alpha$'s and $d=1,2$?

Finally, it is an important open problem to prove bounds on converge rates for densities
that have higher order smoothness (\ie{} $\beta$-H\"older smooth densities).
A related open problem, in the context of of theory of
Euclidean functionals, is stated in~\cite{Koo-Lee2007}.

\section*{Acknowledgements}

This work was supported in part by AICML, AITF (formerly
iCore and AIF), NSERC, the PASCAL2 Network
of Excellence under EC grant no. 216886 and by the Department of Energy 
under grant number DESC0002607. Cs. Szepesv\'ari
is on leave from SZTAKI, Hungary.

\small
\bibliographystyle{plainnat}
\bibliography{biblio}

\appendix
\newpage
\normalsize
\section{Quasi-Additive and Very Strong Euclidean Functionals}
\label{section:quasi-additive-very-strong}

The basic tool to prove convergence properties of our estimators is the theory
of quasi-additive Euclidean functionals developed
by~\cite{Yukich1998,Steele1997,Redmond-Yukich1996,Koo-Lee2007} and others.  We
apply this machinery to the nearest neighbor functional  $L_p$ defined in
equation (\ref{equation:nn-functional}). 

In particular, we use the axiomatic definition of a quasi-additive Euclidean
functional from~\cite{Yukich1998} and the definition of a very strong Euclidean
functional from~\cite{Koo-Lee2007} who add two extra axioms.
We then use the results of \cite{Redmond-Yukich1996} and \cite{Koo-Lee2007}
which hold for these kinds of functionals. These results determine the limit
behavior of the functionals on a set of points chosen i.i.d. from an absolutely
continuous distribution over $\R^d$. As we show in the
following sections, the nearest neighbor functional $L_p$ defined by equation
(\ref{equation:nn-functional}) is a very strong Euclidean functional and thus
both of these results apply to it. 

Technically, a quasi-additive Euclidean functional is a pair of real
non-negative functionals $(L_p(V), L_p^*(V,B))$ where $B \subset \R^d$ is a
$d$-dimensional cube and $V \subset B$ is a finite set of points.  Here, a 
$d$-dimensional \emph{cube} is a set of the form $\prod_{i=1}^d
[a^i,a^i+s]$ where $(a^1, a^2, \dots, a^d) \in \R^d$ is its ``lower-left''
corner and $s > 0$ is its side.  The functional $L_p^*$ is called the
\emph{boundary functional}. The common practice is to neglect $L_p^*$ and refer
to the pair $(L_p(V), L_p^*(V,B))$ simply as $L_p$.  We provide a boundary
functional $L_p^*$ for the nearest neighbor functional $L_p$ in the next
section.

\begin{definition}[Quasi-additive Euclidean functional]
$L_p$ is a quasi-additive Euclidean functional of power $p$ if it satisfies axioms
(A1)--(A7) below.
\end{definition}

\begin{definition}[Very strong Euclidean functional]
$L_p$ is a very strong Euclidean functional of power $p$ if it satisfies axioms
(A1)--(A9) below.
\end{definition}

\paragraph{Axioms.}
For all cubes $B \subseteq \R^d$, any finite $V \subseteq B$, all $\y \in \R^d$, all $t > 0$,
\begin{align}
\tag{A1}
L_p(\emptyset) & = 0\;;  &   L_p^*(\emptyset, B) & = 0\;;
\\
\tag{A2}
L_p(\y + V) & = L_p(V)\;;  &   L_p^*(\y + V,\y + B) & = L_p^*(V,B)\;;
\\
\tag{A3}
L_p(tV) & = t^p L_p(V)\;;  &   L_p^*(tV,tB) & = t^p L_p(V,B)\;;
\\
\tag{A4}
L_p(V) & \ge L^*_p(V,B)\;.
\end{align}
For all $V \subseteq [0,1]^d$ and a partition $\{Q_i ~:~ 1 \le i \le m^d \}$ of $[0,1]^d$ into $m^d$
subcubes of side $1/m$
\begin{align}
\tag{A5}
L_p(V) & \le \sum_{i=1}^{m^d} L_p(V \cap Q_i) + O(m^{d-p})\;,
&
L_p^*(V,[0,1]^d) & \ge \sum_{i=1}^{m^d} L_p^*(V \cap Q_i, [0,1]^d) - O(m^{d-p})\;.
\end{align}
For all finite $V,V' \subseteq [0,1]^d$,
\begin{align}
\tag{A6}
|L_p(V') - L_p(V)| & \le O(|V'\Delta V|^{1-p/d})\;; &
|L_p^*(V',[0,1]^d) - L_p^*(V,[0,1]^d)| & \le O(|V'\Delta V|^{1-p/d})
\end{align}
For a set $\U_n$ of $n$ points drawn i.i.d. from the uniform distribution over $[0,1]^d$,
\begin{align}
\tag{A7}
|\Exp L_p(\U_n) - \Exp L_p^*(\U_n, [0,1]^d)| & \le o(n^{1-p/d})\;;
\\
\tag{A8}
|\Exp L_p(\U_n) - \Exp L^*_p(\U_n, [0,1]^d)| & \le O(\max(n^{1-p/d-1/d},1)) \; ;
\\
\tag{A9}
|\Exp L_p(\U_n) - \Exp L_p(\U_{n+1})| & \le O(n^{-p/d}) \; .
\end{align}

Axiom (A2) is translation invariance, axiom (A3) is scaling.  First part of
(A5) is subadditivity of $L_p$ and second part is super-additivity of $L^*_p$.
Axiom (A6) is smoothness and we call (A7) quasi-additivity. Axiom (A8) is a
strengthening of (A7) with an explicit rate. Axiom (A9) is the add-one bound.
The axioms in \cite{Koo-Lee2007} are slightly different, however it is a routine
to check that they are implied by our set of axioms.

We will use two fundamental results about Euclidean functionals.
The first is \cite[Theorem~2.2]{Redmond-Yukich1996} and the second is essentially 
\cite[Theorem~4]{Koo-Lee2007}.

\begin{theorem}[Redmond-Yukich]
\label{theorem:Redmond-Yukich}
Let $L_p$ be \emph{quasi-additive Euclidean functional} of power $0 < p < d$.
Let $\V_n$ consist of $n$ points drawn i.i.d. from an absolutely continuous distribution
over $[0,1]^d$ with common probability density function $f:[0,1]^d \to \R$. Then,
$$
\lim_{n \to \infty} \frac{L_p(\V_n)}{n^{1-p/d}} = \gamma \int_{[0,1]^d} f^{1-p/d}(\x) \ \ud \x
\quad \text{a.s.} \; ,
$$
where $\gamma := \gamma(L_p, d)$ is a constant depending only on the functional $L_p$ and $d$.
\end{theorem}

\begin{theorem}[Koo-Lee]
\label{theorem:Koo-Lee}
Let $L_p$ be a \emph{very strong Euclidean functional} of power $0 < p < d$.
Let $\V_n$ consist of $n$ points drawn i.i.d. from an absolutely distribution
over $[0,1]^d$ with common probability density
function $f:[0,1]^d \to \R$. If $f$ is Lipschitz~\footnote{Recall that a function $f$ is Lipschitz
if there exists a constant $C > 0$ such that $|f(x)-f(y)| \le C\|x-y\|$ for all $x,y$ in the domain of $f$.},
then
$$
\left| \frac{\Exp L_p(\V_n)}{n^{1-p/d}} \  - \ \gamma \int_{[0,1]^d} f^{1-p/d}(\x) \ \ud \x  \right|
\le
\begin{cases}
O\left( n^{-\frac{d-p}{d(2d-p)}} \right), & \text{if $0 < p < d-1$}\; ; \\
O\left( n^{-\frac{d-p}{d(d+1)}} \right), & \text{if $d-1 \le p < d$} \; ,
\end{cases}
$$
where $\gamma$ is the constant from Theorem~\ref{theorem:Redmond-Yukich}.
\end{theorem}

Theorem~\ref{theorem:Koo-Lee} differs from its original statement
\cite[Theorem~4]{Koo-Lee2007} in two ways. First, our version is restricted to
Lipschitz densities.  Koo and Lee prove a generalization of
Theorem~\ref{theorem:Koo-Lee} for $\beta$-H\"older smooth density functions.
The coefficient $\beta$ then appears in the exponent of $n$ in the rate.
However, their result holds only for $\beta$ in the interval $(0,1]$ which does
not make it very interesting.  The case $\beta=1$ corresponds to
Lipschitz densities and is perhaps the most important in this range.  Second,
Theorem~\ref{theorem:Koo-Lee} has slight improvement in the rate. Koo and Lee
have an extraneous $\log(n)$ factor which we remove by ``correcting'' their
axiom (A8).  

In the next section, we prove that the nearest neighbor functional $L_p$
defined by (\ref{equation:nn-functional}) is a very strong Euclidean
functional. First, in section~\ref{section:nn-boundary-functional}, we provide
a boundary functional $L_p^*$ for $L_p$.  Then, in
section~\ref{section:verification-of-axioms}, we verify that $(L_p, L_p^*)$
satisfy axioms (A1)--(A9). Once the verification is done,
Theorem~\ref{theorem:constant-gamma} follows from
Theorem~\ref{theorem:Redmond-Yukich}.  

Theorem~\ref{thm:mainentropy} will follow from Theorem~\ref{theorem:Koo-Lee}
and a concentration result.  We prove the concentration result in
Section~\ref{section:concentration} and finish that section with the proof of
Theorem~\ref{thm:mainentropy}.  Proof of Theorem~\ref{thm:maininformation}
requires more work---we need to deal with the effect of empirical copula
transformation.  We handle this in Section~\ref{section:copula-MI-appendix} by
employing the classical Kiefer-Dvoretzky-Wolfowitz theorem.

\section{The Boundary Functional $L_p^*$}
\label{section:nn-boundary-functional}

We start by constructing the nearest neighbor boundary functional $L_p^*$.  For
that we will need to introduce an auxiliary graph, which we call the
nearest-neighbor graph \emph{with boundary}. This graph is related to $NN_S$
and will be useful later. 

Let $B$ be a $d$-dimensional cube, $V \subset B$ be finite, and $S \subset
\N^+$ be non-empty and finite.  We define nearest-neighbor graph with boundary
$NN^*_S(V,B)$ to be a directed graph, with possibly parallel edges, on vertex
set $V \cup \partial B$, where $\partial B$ denotes the boundary of $B$.
Roughly speaking, for every vertex $\x \in V$ and every $i \in S$ there is an
edge to its ``$i$-th nearest-neighbor'' in $V \cup \partial B$.

\begin{figure}[h]
\begin{center}
\subfloat[$NN_S(V)$]{\includegraphics[width=0.45\linewidth]{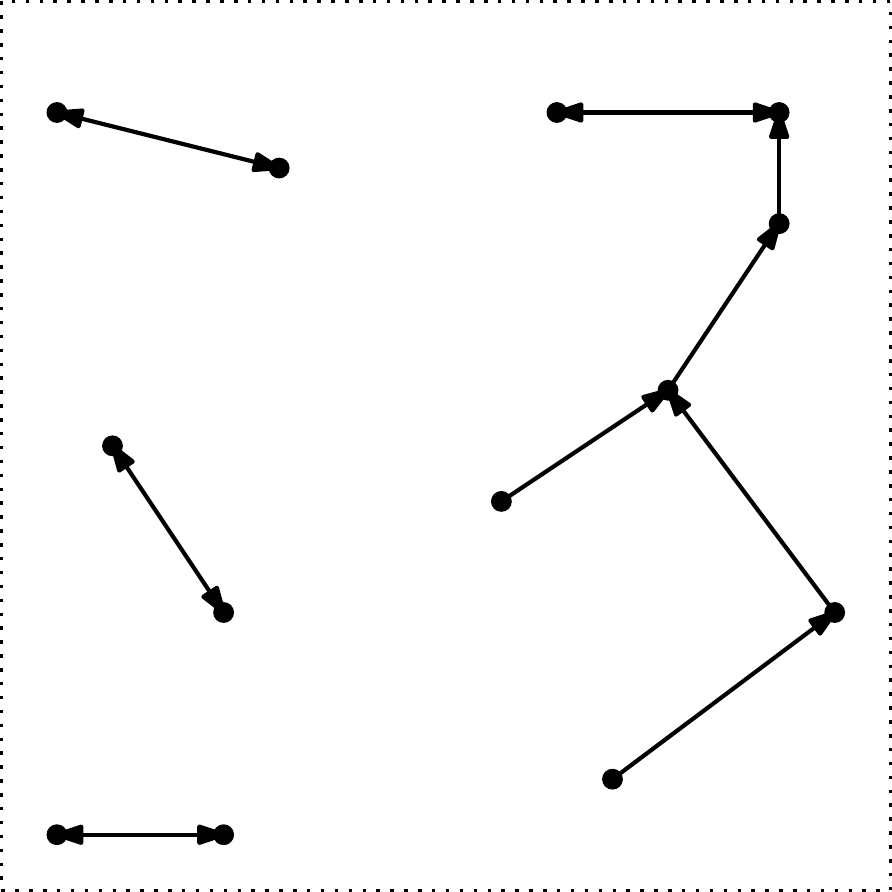}\label{figure:1nn-graph}}
\qquad
\subfloat[$NN_S^*(V,B)$]{\includegraphics[width=0.45\linewidth]{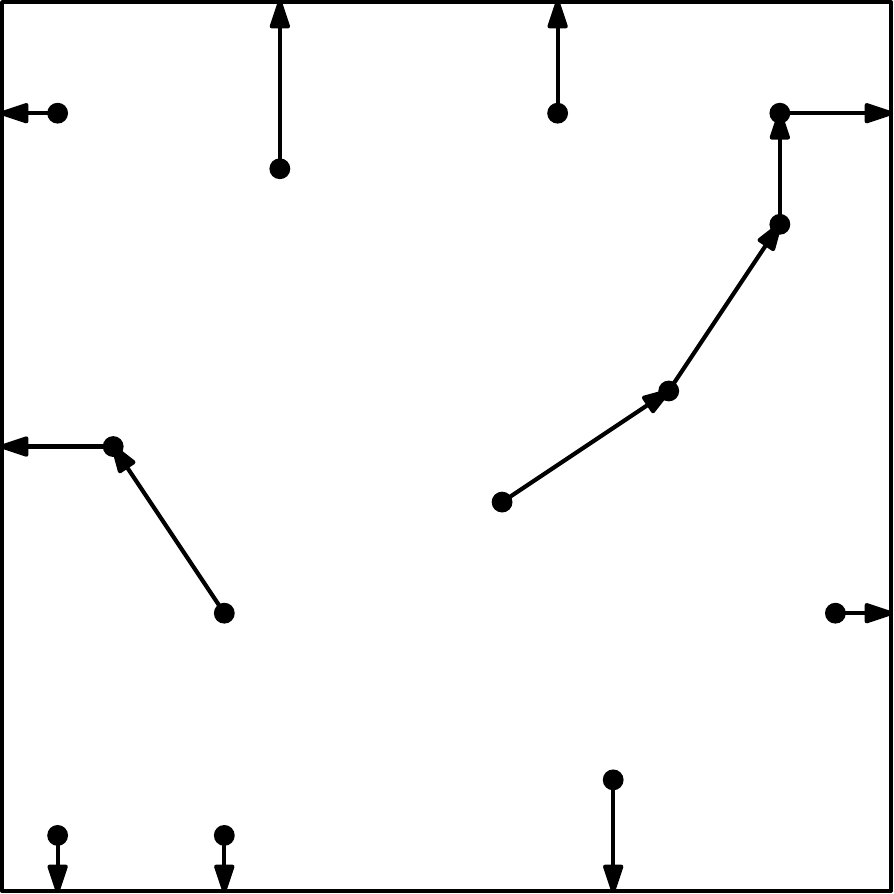}\label{fig:1nn-boundary-graph}}
\caption{Figure (a) shows an example of a nearest neighbor graph $NN_S(V)$ in two dimensions 
and a corresponding boundary nearest neighbor graph $NN_S(V,B)^*$ is shown in Figure (b).
We have used $S=\{1\}$, $B=[0,1]^2$ and a set $V$ consisting of $13$ points in $B$.}
\label{figure:1nn}
\end{center}
\end{figure}

More precisely, we define the edges from $\x \in V$ as follows: Let $\b \in
\partial B$ be the boundary point closest to $\x$. (If there are multiple
boundary points that are the closest to $\x$ we choose one arbitrarily.) If $(\x,\y)
\in E(NN_S(V))$ and $\|\x-\y\| \le \|\x-\b\|$ then $(\x,\y)$ also belongs to
$E(NN^*_S(V,B))$. For each $(\x,\y) \in E(NN_S(V))$ such that $\|\x-\y\| >
\|\x-\b\|$ we create in $NN^*_S(V,B)$ one copy of the edge $(\x,\b)$. In other
words, there is a bijection between edge sets $E(NN_S(V))$ and
$E(NN^*_S(V,B))$. An example of a graph $NN_S(V)$ and a corresponding graph $NN_S^*(V)$
are shown in Figure~\ref{figure:1nn}.

Analogously, we define $L^*_p(V,B)$ as the sum of $p$-powered edges of $NN^*_S(V,B)$. Formally,
\begin{equation}
\label{equation:boundary-nn-functional}
L^*_p(V,B) = \sum_{(\x,\y) \in E(NN_S^*(V,B))} \|\x-\y\|^p \; .
\end{equation}
We will need some basic geometric properties of $NN^*_S(V,B)$ and $L^*_p(V,B)$.  By
construction, the edges of $NN^*_S(V,B)$ are shorter than the corresponding edges
of $NN_S(V)$. As an immediate consequence we get the following proposition.

\begin{proposition}[Upper Bound]
\label{proposition:boundary-functional-bound}
For any cube $B$, any $p \ge 0$ and any finite set $V \subset B$,  $L^*_p(V,B) \le L_p(V)$.
\end{proposition}

\section{Verification of Axioms (A1)--(A9) for $(L_p, L_p^*)$}
\label{section:verification-of-axioms}

It is easy to see that the nearest neighbor functional $L_p$ and its boundary
functional $L_p^*$ satisfy axioms (A1)--(A3). Axiom (A4) is verified by
Proposition~\ref{proposition:boundary-functional-bound}. It thus remains to
verify axioms (A5)--(A9) which we do in subsections~\ref{section:smoothness},
\ref{section:subadditivity-superadditivity} and \ref{section:uniform-distribution}.
We start with two simple lemmas.

\begin{lemma}[In-Degree]
\label{lemma:in-degree}
For any finite $V \subseteq \R^d$ the in-degree of any vertex
in $NN_S(V)$ is $O(1)$.
\end{lemma}

\begin{proof}
Fix a vertex $\x \in V$. We show that the in-degree of $\x$ is bounded by some constant
that depends only on $d$ and $k = \max S$. For any unit vector $\u \in \R^d$ we consider
the convex open cone $C(\x,\u)$ with apex at $\x$, rotationally symmetric about its axis $\u$
and angle $30^\circ$:
$$
Q(\x,\u) = \left\{ \y \in \R^d ~:~ \u \cdot (\y-\x) < \frac{\sqrt{3}}{2}\|\u-\x\| \right\} \; .
$$
As it is well known, $\R^d \setminus \{\x\}$ can be written as a union of
finitely many, possibly overlapping, cones $Q(\x,\u_1)$, $Q(\x,\u_2)$, $\dots$, $Q(\x,\u_B)$,
where $B$ depends only on the dimension $d$. We show that the in-degree of $\x$ is at most $kB$.

Suppose, by contradiction, that the in-degree of $\x$ is larger than $kB$. Then, by pigeonhole
principle, there is a cone $Q(\x,\u)$ containing $k+1$ vertices of the graph
with an incoming edge to $\x$. Denote these vertices $\y_1, \y_2, \dots, \y_{k+1}$
and assume that they are indexed so that $\|\x-\y_1\| \le \|\x-\y_2\| \le \dots \le \|\x-\y_{k+1}\|$.

By a simple calculation, we can verify that $\|\x-\y_{k+1}\| > \|\y_{i} - \y_{k+1}\|$
for all $1 \le i \le k$. Indeed, by the law of cosines
\begin{multline*}
\|\y_i - \y_{k+1} \|^2 = \|\x-\y_i\|^2 + \|\x-\y_{k+1}\|^2 - 2(\x-\y_i) \cdot (\x-\y_{k+1}) \\
< \|\x-\y_i\|^2 + \|\x-\y_{k+1}\|^2 - \|\x-\y_i\| \|\x-\y_{k+1}\| \le \|\x-\y_{k+1}\|^2 \; ,
\end{multline*}
where the sharp inequality follows from that $\y_{k+1},\y_i \in Q(\x,\u)$ and so the angle between vectors
$(\x-\y_i)$ and $(\x-\y_{k+1})$ is strictly less than $60^\circ$, and the second
inequality follows from $\|\x-\y_i\| \le \|\x-\y_{k+1}\|$.
Thus, $\x$ cannot be among the $k$ nearest-neighbors of $\y_{k+1}$ which contradicts
the existence of the edge $(\y_{k+1},\x)$.
\end{proof}

\begin{lemma}[Growth Bound]
\label{lemma:growth-bound}
For any $p \ge 0$ and finite $V \subset [0,1]^d$, $L_p(V) \le O(\max(|V|^{1-p/d}, 1))$.
\end{lemma}

\begin{proof}
An elegant way to prove the lemma is with the use of space-filling curves.\footnote{There is an elementary proof, too, based on a discretization argument.
However, this proof introduces an extraneous logarithmic factor when $p=d$.} Since \cite{Peano1890} and \cite{Hilbert1891},
it is known that there exists a continuous function $\bm{\psi}$ from the unit interval $[0,1]$ \textbf{onto} the cube $[0,1]^d$ (\ie{} a surjection).
For obvious reason $\bm{\psi}$ is called a space-filling curve. Moreover, there are space-filling
curves which are  $(1/d)$-H\"older; see \cite{Milne1980}. In other words, we can assume that there exists a
constant $C > 0$ such that
\begin{align}
\label{equation:space-filling-holder}
\|\bm{\psi}(x) - \bm{\psi}(y)\| & \le C|x-y|^{1/d} & \forall x,y \in [0,1].
\end{align}
Since $\bm{\psi}$ is a surjective function we can consider a right inverse $\bm{\psi}^{-1}:[0,1]^d \to
[0,1]$ \ie{} a function such that $\bm{\psi}(\bm{\psi}^{-1}(x)) = x$ and we let $W =
\bm{\psi}^{-1}(V)$.  Let $0 \le w_1 < w_2 < \dots < w_{|V|} \le 1$
be the points of $W$ sorted in the increasing order. We construct a ``nearest
neighbor'' graph $G$ on $W$.  For every $1 \le j \le |V|$ and every $i
\in S$ we create a directed edge $(w_j,w_{j+i})$, where the addition $i+j$ is
taken modulo $|V|$. It is not hard to see that the total length of the edges of
$G$ is
\begin{equation} \label{equation:ast}
\sum_{(x,y) \in E(G)} |x-y| \le O(k^2) =  O(1)
\end{equation}
To see more clearly why~\eqref{equation:ast} holds, note that every
line segment $[w_i, w_{i+1}]$, $1 \le i < |V|$ belongs to at most $O(k^2)$ edges
and the total length of the line segments is $\sum_{i=1}^{|V|-1} (w_{i+1} - w_i)
\le 1$.

Let $H$ be a graph on $V \subset[0,1]^d$ isomorphic to $G$, where for each edge $(w_i, w_j) \in
E(G)$ there is a corresponding edge $(\bm{\psi}(w_i), \bm{\psi}(w_j)) \in E(H)$. By the
construction of $H$
\begin{equation}
\label{equation:dagger}
L_p(V) \le \sum_{(\x,\y) \in E(H)} \|\x-\y\|^p \ = 
\sum_{(x,y) \in E(G)} \!\!\!\! \|\bm{\psi}(x)-\bm{\psi}(y)\|^p  \; .
\end{equation}
H\"older property of $\bm{\psi}$ implies that
\begin{equation}
\label{equation:ddagger}
\sum_{(x,y) \in E(G)} \!\!\!\! \|\bm{\psi}(x)-\bm{\psi}(y)\|^p 
\ \le \ C \!\!\!\! \sum_{(x,y) \in E(G)} \!\!\!\! |x-y|^{p/d} \; .
\end{equation}

If $p \ge d$ then $|x-y|^{p/d} \le |x-y|$ since $|x-y| \in [0,1]$ and thus
$$
\sum_{(x,y) \in E(G)} |x-y|^{p/d} \le \sum_{(x,y) \in E(G)} |x-y| \; .
$$
Chaining the last inequality with \eqref{equation:dagger}, \eqref{equation:ddagger} and \eqref{equation:ast} we obtain that $L_p(V) \le O(1)$ for $p \ge d$.

If $0 < p < d$ we use the inequality between arithmetic and $(p/d)$-mean. It states that
for positive numbers $a_1, a_2, \dots, a_n$
$$
\left(\frac{\sum_{i=1}^n a_i^{p/d}}{n}\right)^{d/p} \le \frac{\sum_{i=1}^n a_i}{n}
\qquad
\text{or equivalently}
\qquad
\sum_{i=1}^n a_i^{p/d} \le n^{1-p/d} \left( \sum_{i=1}^n a_i \right)^{p/d} \; .
$$
In our case $a_i$'s are the edge length of $G$ and $n \le k|V|$, and we have
$$
\sum_{(x,y) \in E(G)} |x-y|^{p/d} \le (k|V|)^{1-p/d} \left( \sum_{(x,y) \in E(G)} |x-y| \right)^{p/d} \; .
$$
Combining the last inequality with \eqref{equation:dagger}, \eqref{equation:ddagger} and \eqref{equation:ast} we get that $L_p(V) \le O(|V|^{1-p/d})$ for $0 < p < d$.

Finally, for $p=0$, $L_p(V) \le k|V| = O(|V|)$.
\end{proof}

\subsection{Smoothness}
\label{section:smoothness}
In this section, we verify axiom (A6).

\begin{lemma}[Smoothness of $L_p$]
\label{lemma:smoothness}
For $p \ge 0$ and finite disjoint $V,V' \subset [0,1]^d$, $|L_p(V' \cup V) - L_p(V')| \le O(\max(|V|^{1-p/d}, 1))$.
\end{lemma}

\begin{proof}
For $p \ge d$ the lemma trivially follows from the growth bound $L_p(V') = O(1)$, $L_p(V' \cup V) = O(1)$.
For $0 \le p < d$, we need to prove two inequalities:
\begin{align*}
L_p(V' \cup V) & \le L_p(V') + O(|V|^{1-p/d})
&  \text{and} \qquad
L_p(V') & \le L_p(V' \cup V) + O(|V|^{1-p/d}) \; .
\end{align*}

We start with the first inequality. We use the obvious property of $L_p$ that
$L_p(V' \cup V) \le L_p(V') + L_p(V) + O(1)$. Combined with the growth bound~(Lemma \ref{lemma:growth-bound})
for $V$ we get
$$
L_p(V' \cup V) \le L_p(V') + L_p(V) + O(1) \le L_p(V') + O(|V|^{1-p/d}) + O(1) \le L_p(V') + O(|V|^{1-p/d}) \; .
$$

The second inequality is a bit more tricky to prove. We introduce a generalized
nearest-neighbor graph $NN_S(W,W')$ for any pair of finite sets $W,W'$ such that $W \subseteq W' \subset \R^d$.
We define $NN_S(W,W')$ as the subgraph of $NN_S(W')$ where all edges from $W' \setminus W$ are deleted.
Similarly, we define $L_p(W,W')$ as the sum $p$-powered lengths of edges of $NN_S(W,W')$:
$$
L_p(W,W') = \sum_{(x,y) \in E(NN_S(W,W'))} \|\x-\y\|^p \; .
$$
We will use two obvious properties of $L_p(W,W')$ valid for any finite $W \subseteq W' \subset \R^d$:
\begin{align}
\label{equation:two-properties}
L_p(W,W) & = L_p(W) & \text{and} \qquad L_p(W,W') & \le L_p(W) + O(1) \; .
\end{align}
Let $U \subseteq V'$
be the set of vertices $\x$ such that in $NN_S(V' \cup V)$ there exists an edge from $\x$
to a vertex $V$. Using the two observations and the growth bound we have
\begin{align*}
L_p(V') = L_p(V',V') = L_p(U,V') + L_p(V' \setminus U, V')
& \le L_p(U) + O(1) + L_p(V' \setminus U, V') \\
& \le O(|U|^{1-p/d}) + L_p(V' \setminus U, V') \; .
\end{align*}
The term is $L_p(V' \setminus U,V')$ can be upper bounded by $L_p(V' \cup V)$
since by the choice of $U$ the graph $NN_S(V' \setminus U,V')$ is a subgraph of $NN_S(V' \cup V)$.
The term $O(|U|^{1-p/d})$ is at most $O(|V|^{1-p/d})$ since $|U|$ is upper bounded by the number of edges
of $NN_S(V' \cup V)$ ending in $V$ and, in turn, the number of these edges is by the in-degree lemma at most $O(|V|)$.
\end{proof}

\begin{corollary}[Smoothness of $L_p$]
\label{corollary:smoothness}
For $p \ge 0$ and finite $V,V' \subset [0,1]^d$, $$|L_p(V') - L_p(V)| \le O(\max(|V'\Delta V|^{1-p/d},1)),$$
where $V' \Delta V$ denotes the symmetric difference.
\end{corollary}

\begin{proof}
Applying the previous lemma twice
\begin{align*}
|L_p(V') - L_p(V)|
& \le |L_p(V') - L_p(V' \cup V)|  + |L_p(V' \cup V) - L_p(V)| \\
& = |L_p(V') - L_p(V' \cup (V \setminus V'))| + |L_p(V \cup (V' \setminus V)) - L_p(V)| \\
& \le O(\max(|V \setminus V'|^{1-p/d},1)) + O(\max(|V' \setminus V|^{1-p/d},1)) \\
& = O(\max(|V' \Delta V|^{1-p/d},1)) \; .
\end{align*}
\end{proof}

\begin{lemma}[Smoothness of $L_p^*$]
\label{lemma:boundary-smoothness}
For $p \ge 0$ and finite disjoint $V,V' \subset [0,1]^d$,
$$
|L^*_p(V' \cup V, [0,1]^d) - L^*_p(V', [0,1]^d)| \le O(\max(|V|^{1-p/d}, 1)) \; .
$$
\end{lemma}

\begin{proof}
The proof of the lemma is identical to the proof of Lemma~\ref{lemma:smoothness}
if we replace $L_p(\cdot)$ by $L_p^*(\cdot, [0,1]^d)$, $NN_S^*(\cdot)$
by $NN_S(\cdot, [0,1]^d)$, $L_p(\cdot, \cdot)$ by $L_p^*(\cdot, \cdot, [0,1]^d)$
and $NN_S(\cdot, \cdot)$ by $NN_S^*(\cdot, \cdot, [0,1]^d)$.
We, of course, need to explain what $NN_S^*(V,W,[0,1]^d)$ and $L_p^*(V,W,[0,1])$ mean.
For $V \subseteq W$, we define $NN_S^*(V,W,[0,1]^d)$ as the subgraph of $NN_S^*(W,[0,1]^d)$,
where the edges starting in $W \setminus V$ are removed, and $L_p^*(V,W,[0,1]^d)$
is the sum the $p$-th powers of Euclidean lengths of edges of $NN_S^*(V,W,[0,1]^d)$.
\end{proof}

\begin{corollary}[Smoothness of $L_p^*$]
\label{corollary:boundary-smoothness}
For $p \ge 0$ and finite $V,V' \subset [0,1]^d$,
$$
|L_p^*(V',[0,1]^d) - L_p^*(V,[0,1]^d)| \le O(\max(|V'\Delta V|^{1-p/d},1)) \; ,
$$
where $V' \Delta V$ denotes the symmetric difference.
\end{corollary}

\begin{proof}
The corollary is proved in exactly the same way as Corollary~\ref{corollary:smoothness},
where $L_p(\cdot)$ is replaced by $L_p^*(\cdot,[0,1]^d)$.
\end{proof}

\subsection{Subadditivity and Superadditivity}
\label{section:subadditivity-superadditivity}

In this section, we verify axiom (A5).

\begin{lemma}[Subadditivity]
\label{lemma:subadditivity}
Let $p \ge 0$. For $m \in \N^+$ consider
the partition $\{ Q_i ~:~ 1 \le i \le m^d \}$ of the cube $[0,1]^d$ into $m^d$ disjoint subcubes\footnote{In order
the subcubes to be pairwise disjoint, most of them need to be semi-open and some of them closed.} of side $1/m$. For any finite $V \subset [0,1]^d$,
\begin{equation}
\label{equation:subadditivity}
L_p(V) \le \sum_{i=1}^{m^d} L_p(V \cap Q_i) + O(\max(m^{d-p},1)) \; .
\end{equation}
\end{lemma}

\begin{proof}
Consider a subcube $Q_i$ which contains at least $k+1$ points.
Using the ``$L_p(W,W')$ notation'' from the proof of Lemma~\ref{lemma:smoothness}
$$
L_p(V \cap Q_i, V) \le L_p(V \cap Q_i, V \cap Q_i) = L_p(V \cap Q_i) \; .
$$
Let $R$ be the union subcubes that contain at most $k$ points. Clearly $|V \cap R| \le k m^d$.
Then
\begin{align*}
L_p(V) 
& = L_p(V,V) \\
& = L_p(V \cap R,V) + \hspace*{-3mm}\sum_{\substack{1 \le i \le m^d \\ |V \cap Q_i| \ge k+1}} \hspace*{-3mm}L_p(V \cap Q_i, V) \\
& \le L_p(V \cap R) + O(1) + \sum_{i=1}^{m^d} L_p(V \cap Q_i) \; ,
\end{align*}
where we have used the second part of (\ref{equation:two-properties}). The proof
is finished by applying the growth bound $L_p(V \cap R) \le O(\max(|V \cap R|^{1-p/d},1)) \le O(\max(m^{d-p},1))$.
\end{proof}
\bigskip

\begin{lemma}[Superadditivity of $L_p^*$]
\label{lemma:superadditivity}
Let $p \ge 0$. For $m \in \N^+$ consider a partition $\{ Q_i ~:~ 1 \le i \le m\}$
of $[0,1]^d$ into $m^d$ disjoint subcubes of side $1/m$. For any finite $V \subset [0,1]^d$,
$$
\sum_{i=1}^{m^d} L^*_p(V \cap Q_i,Q_i) \le L^*_p(V,[0,1]^d) \; .
$$
\end{lemma}

\begin{proof}
We construct a new graph $\hat G$ by modifying the graph $NN^*_S(V,[0,1]^d)$.
Consider any edge $(\x,\y)$ such that $\x \in Q_i$ and $\y \not \in Q_i$ for some
$1 \le i \le m^d$.  Let $\z$ be the point where $\partial Q_i$ and the line
segment from $\x$ to $\y$ intersect.  In $\hat G$, we replace $(\x,\y)$ by $(\x,\z)$.
Note that the all edges of $\hat G$ lie completely in one of the subcubes $Q_i$
and they are shorter or equal to the corresponding edges in $NN^*_S(V,[0,1]^d)$.

Let $\hat L_{i,p}$ be the sum of $p$-th powers of the Euclidean length of the edges
of $\hat G$  lying in $Q_i$. Since edges in $\hat G$ are shorter than in $NN^*_S(V,[0,1]^d)$,
 $\sum_{i=1}^{m^d} \hat L_{p,i} \le L^*_p(V,[0,1]^d)$. To finish the proof it remains
to show that $L^*_p(V \cap Q_i, Q_i) \le \hat L_{i,p}$ for all $1 \le i \le m^d$.

For any edge $(\x,\z)$ in $\hat G$ from $\x \in V \cap Q_i$ to $\z \in \partial Q_i$,
the point $\z \in \partial Q_i$ is not necessarily the closest to $\x$. Therefore,
any edge in $NN^*_S(V \cap Q_i, Q_i)$ is shorter than the corresponding edge in $\hat G$.
\end{proof}

\subsection{Uniformly Distributed Points}
\label{section:uniform-distribution}

Axiom (A7) is a direct consequence of axiom (A8). Hence, we are left with
verifying axioms (A8) and (A9). In this section, $\U_n$ denotes a set of $n$
points chosen independently uniformly at random from $[0,1]^d$.

\begin{lemma}[Average Edge Length]
\label{lemma:average-edge-length}
Assume $\X_1, \X_2, \dots, \X_n$ are chosen i.i.d. uniformly at random from $[0,1]^d$.
Let $k$ be a fixed positive integer. Let $Z$ be the distance from $\X_1$ to $k$-th
nearest-neighbor in $\{\X_2, \X_3, \dots, \X_n\}$.
For any $p \ge 0$,
$$
\Exp[Z^p~|~\X_1] \le O(n^{-p/d})\;.
$$
\end{lemma}

\begin{proof}
We denote by $B(\x,r) = \{ \y \in \R^d ~:~ \|\x-\y\| \le r \}$ the ball of radius of $r \ge 0$ centered at a point $\x \in \R^d$.
Since $Z$ lies in the interval $[0,\sqrt{d}]$ is non-negative,
\begin{align*}
\Exp[Z^p~|~\X_1]
& = \int_{0}^\infty \Pr[Z^p > t~|~\X_1] \ \ud t \\
& = p \int_{0}^{\sqrt{d}} u^{p-1} \Pr[Z > u~|~\X_1] \ud u \\
& = p \int_{0}^{\sqrt{d}} u^{p-1} \Pr[ |\{\X_2, \X_3, \dots, \X_n\} \cap B(\X_1,u)| < k  ~|~ \X_1 ] \ \ud u \\
& = p \int_{0}^{\sqrt{d}} \sum_{j=0}^{k-1} \binom{n-1}{j} u^{p-1} \left[ \Vol(B(\X_1,u) \cap [0,1]^d) \right]^j \\
& \qquad \cdot \left[ 1-\Vol(B(\X_1,u) \cap [0,1]^d) \right]^{n-1-j} \  \ud u \\
& \le p \int_{0}^{2\sqrt{d}} \sum_{j=0}^{k-1} \binom{n-1}{j} u^{p-1} \left[ \Vol(\X_1, u) \right]^j
\left[1-\left(\frac{u}{2\sqrt{d}}\right)^d \right]^{n-1-j} \ \ud u\; .
\end{align*}
The last inequality follows from the obvious bound $\Vol(B(\X_1,u) \cap [0,1]^d) \le \Vol(B(\X_1, u))$
and that for $u \in [0,\sqrt{d}]$ the intersection
$B(\X_1,u) \cap [0,1]^d$ contains a cube of side at least $\frac{u}{2\sqrt{d}}$.
To simplify this complicated integral, we note that $\Vol(B(\X_1, u)) = \Vol(B(\X_1,1))u^d$ and make
substitution $s = (\frac{u}{2\sqrt{d}})^d$.  The last integral can be bounded by a
constant multiple of
$$
\sum_{j=0}^{k-1} \binom{n-1}{j} \int_0^1 s^{p/d+j-1} (1-s)^{n-1-j} \ \ud s \; .
$$
Since $\binom{n-1}{j} = O(n^j)$ and the sum consists of only constant number of terms,
it remains to show that the inner integral is $O(n^{-p/d-j})$.
We can express the inner integral using the gamma function. Then, we use the
asymptotic relation $\binom{n}{\epsilon} = \Theta(n^\epsilon)$
for generalized binomial coefficients $\binom{a}{b} = \frac{\Gamma(a+1)}{\Gamma(b+1)\Gamma(a-b+1)}$
to upper-bound the result:
\begin{align*}
\int_0^1 s^{p/d+j-1} (1-s)^{n-1-j} \ \ud s
& = \frac{\Gamma(p/d+j)\Gamma(n-j)}{\Gamma(n+p/d)} \\
& = \frac{1}{\displaystyle (p/d+j)\binom{n+p/d-1}{p/d+j}} \\
& = O(n^{-p/d-j})  \; . \qedhere
\end{align*}
\end{proof}

\begin{lemma}[Add-One Bound]
\label{lemma:add-one-bound}
For any $p \ge 0$, $|\Exp[L_p(\U_n)] - \Exp[L_p(\U_{n+1})]| \le O(n^{-p/d})$.
\end{lemma}

\begin{proof}
Let $\X_1,\X_2, \dots,\X_n,\X_{n+1}$ be i.i.d. points from the uniform distribution over $[0,1]^d$.
We couple $\U_n$ and $\U_{n+1}$ in the obvious way
$\U_n = \{\X_1, \X_2, \dots, \X_n\}$ and $\U_{n+1} = \{\X_1, \X_2, \dots, \X_{n+1}\}$.
Let $Z$ be the distance from $\X_{n+1}$ to $k$-th closest neighbor in $\U_n$.
The inequality
$$
L_p(\U_{n+1}) \le L_p(\U_n) + |S|Z^p
$$
holds since $|S|Z^p$ accounts for the edges from $\X_{n+1}$ and
since the edges from $\U_n$ are shorter (or equal) in $NN_S(\U_{n+1})$
than the corresponding edges in $\U_{n+1}$. Taking expectations and using  Lemma~\ref{lemma:average-edge-length}
we get
$$
\Exp[L_p(\U_{n+1})] \le \Exp[L_p(\U_n)] + O(n^{-p/d}) \; .
$$

To show the other direction of the inequality, let $Z_i$ be the distance
from $\X_i$ its $(k+1)$-the nearest point in $\U_{n+1}$. (Recall that $k = \max S$.)
Let $N(j) = \{ \X_i ~:~ (\X_i,\X_j) \in E(NN_S(\U_{n+1}))\}$ be the incoming neighborhood of $\X_j$.
Now if we remove $\X_j$ from $NN_S(V)$, the vertices in $N(j)$ lose $\X_j$
as their neighbor and they need to be connected to a new neighbor in $\U_{n+1} \setminus \{\X_j\}$.
This neighbor is not farther than their $(k+1)$-th nearest-neighbor in $\U_{n+1}$.
Therefore,
$$
L_p(\U_{n+1} \setminus \{\X_j\}) \le L_p(\U_{n+1}) + \sum_{\X_i \in N(j)} Z_i^p \; .
$$
Summing over all $j=1,2,\dots,n+1$ we have
$$
\sum_{j=1}^{n+1} L_p(\U_{n+1} \setminus \{\X_j\}) \le (n+1)L_p(\U_{n+1}) + \sum_{j=1}^{n+1} \sum_{\X_i \in N(j)} Z_i^p\; .
$$
The double sum on the right hand side is simply the sum over all edges of $NN_S(\U_{n+1})$ and so we can write
$$
\sum_{j=1}^{n+1} L_p(\U_{n+1} \setminus \{\X_j\}) \le (n+1) L_p(\U_{n+1}) + |S|\sum_{i=1}^{n+1} Z_i^p \; .
$$
Taking expectations and using Lemma~\ref{lemma:average-edge-length} to bound $\Exp[Z_i^p]$ we arrive at
$$
(n+1) \Exp[L_p(\U_n)] \le (n+1)\Exp[L_p(\U_{n+1})] + (n+1)O(n^{-p/d}) \; .
$$
The proof is finished by dividing through by $(n+1)$.
\end{proof}

\begin{lemma}[Quasi-additivity] \label{lemma:quasi-additivity}
$0 \le \Exp[L_p(\U_n)] - \Exp[L^*_p(\U_n, [0,1]^d)] \le O(\max(n^{1-p/d-1/d}, 1))$ for any $p \ge 0$.
\end{lemma}

\begin{proof}
The first inequality follows from Proposition~\ref{proposition:boundary-functional-bound}
by taking expectation. The proof of the second inequality is much more involved.
Consider the (random) subset of points $\hat \U_n \subseteq \U_n$ which are connected
to the boundary in $NN^*_S(\U_n,[0,1]^d)$ by at least one edge. We use the notation
$L_p(W,W')$ for any $W \subseteq W'$ and its two properties expressed by Eq.~(\ref{equation:two-properties})
and a third obvious property $L_p(W,W') \le L_p(W')$. We have
$$
L_p(\U_n) = L_p(\U_n, \U_n) = L_p(\widehat \U_n, \U_n) + L_p(\U_n \setminus \widehat \U_n, \U_n)
\le L_p(\widehat \U_n) + O(1) + L^*_p(\U_n, [0,1]^d) \; ,
$$
where in the last step we have used that $L_p(\U_n \setminus \widehat \U_n, \U_n) \le L^*_p(\U_n, [0,1]^d)$
which holds since the edges from vertices $\U_n \setminus \widehat \U_n$ are the same
in both graphs $NN_S(\U_n)$ and $NN^*_S(\U_n, [0,1]^d)$. If we take expectation, we get
$$
\Exp[L_p(\U_n)] - \Exp[L^*_p(\U_n, [0,1]^d)] \le \Exp[L_p(\widehat \U_n)] + O(1)
$$
and we see that we are left to show that $\Exp[L_p(\widehat \U_n)] \le O(\max(n^{1-p/d-1/d},1))$.
In order to do that, we start by showing that
\begin{equation}
\label{equation:boundary-points}
\Exp[|\widehat \U_n|] \le O(n^{1-1/d})  \; .
\end{equation}
Consider the cube $B = [n^{-1/d},1-n^{-1/d}]^d$. We bound $\Exp[|\widehat \U_n
\cap B|]$ and $\Exp[|\widehat \U_n \cap ([0,1]^d \setminus B)|]$ separately.
The latter is easily bounded by $O(n^{1-1/d})$ since there are $n$ points and the
probability that a point lies in  $[0,1]^d \setminus B$ is $\Vol([0,1]^d
\setminus B) \le O(n^{-1/d})$.  We now bound $|\widehat \U_n \cap B|$. Consider
a face of $F$. Partition $B$ into $m = \Theta(n^{1-1/d})$ rectangles $R_1, R_2,
\dots, R_m$ such that the perpendicular projection of any rectangle $R_i$, $1
\le i \le m$, on $F$ has diameter at most $n^{-1/d}$ and its
$(d-1)$-dimensional volume is $\Theta(n^{1/d-1})$; see
Figure~\ref{figure:rectangles}.  It is not hard to see that, in $\U_n \cap
R_i$, only the $k$ closest points to $F$ can be connected to $F$ by an edge in
$NN^*_S(\U_n, [0,1]^d)$. There are $2d$ faces and $m$ rectangles and hence
$|\widehat \U_n \cap B| \le 2dkm = O(n^{1-1/d})$.  We have thus
proved~(\ref{equation:boundary-points}).

\begin{figure}
\centering
\includegraphics{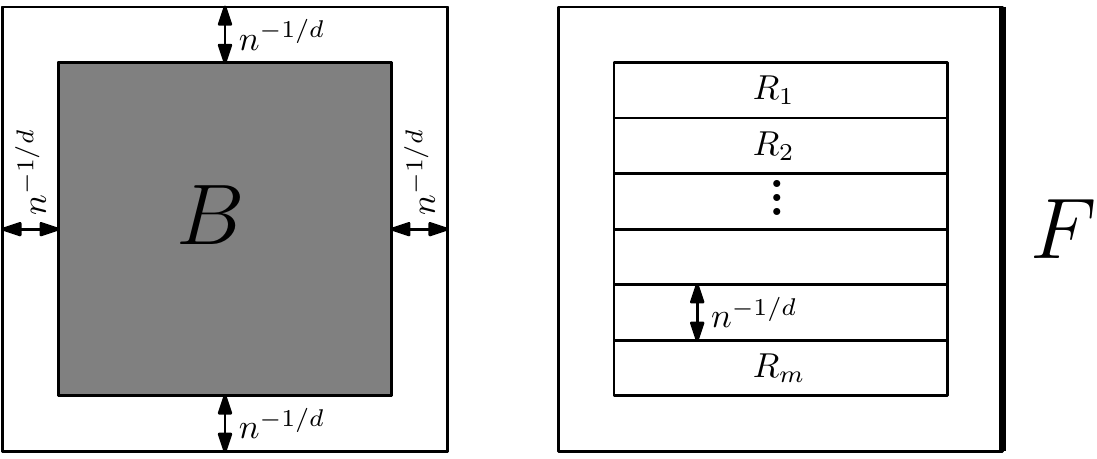}
\caption{\label{figure:rectangles}
The left drawing shows the box $B = [n^{-1/d},1-n^{-1/d}]^d \subset [0,1]^d$ shown in gray.
The right drawing shows partition of $B$ into rectangles $R_1, R_2, \dots, R_m$.
The diameter of the projection of each rectangle $R_i$ on the right side $F$ has diameter (at most) $n^{-1/d}$.
In each rectangle $R_i$ at most $k$ points are connected to $F$ by an edge.
}
\end{figure}

The second key component that we need is that the expected sum of $p$-th powers of
lengths of edges of $NN_S^*(\U_n,[0,1]^d)$ that connect points in $\U_n$ to
$\partial[0,1]^d$ is ``small''. More precisely, for any point $\x \in [0,1]^d$
let $\b_\x \in \partial[0,1]^d$ be the boundary point closest to $\x$.  We show
that
\begin{equation}
\label{equation:boundary-edges}
\Exp\left[ \sum_{\X \in \widehat \U_n} \|\X - \b_\X\|^p \right] \le O(n^{1-p/d-1/d}) \; .
\end{equation}
We decompose the task as
$$
\Exp\left[\sum_{\X \in \widehat \U_n} \|\X - \b_\X\|^p \right]
=
\Exp\left[ \sum_{\X \in \widehat \U_n \cap B} \|\X - \b_\X\|^p \right]
+
\Exp\left[ \sum_{\X \in \widehat \U_n \cap ([0,1]^d \setminus B)} \|\X - \b_\X\|^p \right] \; .
$$
Clearly, the second term is bounded by $n^{-p/d} \Exp[|\widehat \U_n \cap
([0,1]^d \setminus B)|] = O(n^{1-1/d-1/p})$.  To bound the first term, consider
a face $F$ of the cube $[0,1]^d$ and a rectangle $R_i$ in the decomposition of
$B$ into $R_1, R_2, \dots, R_m$ mentioned above. Let $Z$ be the distance of the
$k$-th closest point in $\widehat \U_n \cap R_i$ to $F$. (If $\widehat \U_n \cap R_i$ contains less than
$k$ points, we define $Z$ to be $1-n^{-1/d}$.) Recall that only the
$k$ closest points of $\widehat \U_n \cap R_i$ can be connected to $F$ and this
distance is bounded by $Z$. There are $2d$ faces,  $m=O(n^{1-1/d})$ rectangles
and at most $k$ points in each rectangle connected to a face. If we can show
that $\Exp[Z^p] = O(n^{-p/d})$, we can upper bound the second term by $2dkm \cdot
O(n^{-1/p}) = O(n^{1-p/d-1/d})$ from which (\ref{equation:boundary-edges})
follows.

We now prove that $\Exp[Z^p] = O(n^{-p/d})$. Let $Y=Z-n^{-1/d}$.  Since
$\Exp[Z^p] \le 2^p\Exp[Y^p] + 2^pn^{-p/d}$ it suffices to to show that
$\Exp[Y^p] = O(n^{-p/d})$.  Let $q$ be the $(d-1)$-dimensional volume of the
projection of $R_i$ to $F$. Recall that  $q = \Theta(n^{1/d-1})$.
Since $Y \in [0,1-2n^{-1/d}]$ we have
\begin{align*}
\Exp[Y^p]
& = p \int_0^{1-2n^{-1/d}} t^{p-1} \Pr[Y > t] \ \ud t \\
& = p \int_0^{1-2n^{-1/d}} t^{p-1} \sum_{j=0}^{k-1} \binom{n}{j} (qt)^j (1-qt)^{n-j} \ \ud t
\displaybreak[2]
\\
& \le p q^{-p} \int_0^1 x^{p-1} \sum_{j=0}^{k-1} \binom{n}{j} x^j (1-x)^{n-j} \ \ud x
\\
& = p q^{-p} \sum_{j=0}^{k-1} \binom{n}{j} \frac{\Gamma(p+j)\Gamma(n-j+1)}{\Gamma(n+p+1)}
\\
& = p q^{-p} \sum_{j=0}^{k-1} \frac{1}{(p+j)} \binom{n}{j} \ / \ \binom{n+p}{p+j} \\
& = \Theta(q^{-p}n^{-p}) = \Theta(n^{-p/d})  \; .
\end{align*}

We now use (\ref{equation:boundary-points}) and (\ref{equation:boundary-edges})
to show that $\Exp[L_p(\widehat \U_n)] \le O(\max(n^{1-p/d-1/d},1))$ which will
finish the proof.  For any point $\X \in \widehat \U_n$ consider the point $\b_\X$
lying on the boundary. Let $\widehat \V_n = \{ \b_\X ~:~ \X \in \widehat \U_n \}$
and let $NN_S(\widehat \V_n)$ be its nearest-neighbor graph. Since $\widehat
\V_n$ lies in a union of $(d-1)$-dimensional faces, by the growth bound
$L_p(\widehat \V_n) \le O(\max(|\widehat \V_n|^{1-p/(d-1)},1))$.  Thus, if $0
\le p < d-1$ we use that $x \mapsto x^{1-p/(d-1)}$ is concave and
(\ref{equation:boundary-points}), and we have
\begin{multline*}
\Exp[L_p(\widehat \V_n)]
\le O\left( \Exp\left[ |\widehat \V_n|^{1-p/(d-1)} \right] \right)
= O\left( \Exp\left[ |\widehat \U_n|^{1-p/(d-1)} \right] \right)
\\
\le O\left( \Exp\left[ |\widehat \U_n| \right]^{1-p/(d-1)} \right) \le O(n^{1-1/d})^{1-p/(d-1)}
\le O(n^{1-p/d-1/d})\;.
\end{multline*}
If $p \ge d-1$ then $L_p(\widehat \V_n) = O(1)$. Therefore, for any $p \ge 0$
\begin{equation}
\label{equation:face-graph}
\Exp[L_p(\widehat \V_n)] \le O(\max(n^{1-p/d-1/d}, 1))
\end{equation}
We construct a nearest-neighbor graph $\widehat G$ on $\widehat \U_n$ by
lifting $NN_S(\widehat \V_n)$. For every edge, $(\b_\X,\b_\Y)$ in $NN_S(\widehat
\V_n)$ we create an edge $(\X,\Y)$. Clearly, $L_p(\widehat U_n)$ is at most the
sum of $p$-the powers of the edges lengths of $\widehat G$.  By triangle
inequality, for any $p > 0$
\begin{align*}
\|\X - \Y\|^p 
& \le \left(\|\X - \b_\X\| + \|\b_\X - \b_\Y\| + \|\b_\Y - \Y\| \right)^p \\
& \le 3^p \left( \|\X - \b_\X\|^p + \|\b_\X - \b_\Y\|^p + \|\b_\Y - \Y\|^p \right) \; .
\end{align*}
In-degrees and out-degrees of $\widehat G$ are $O(1)$ and so if we sum over all edges of $(\X,\Y)$ of $\widehat G$ and take expectation, we get
\begin{align*}
\Exp[L_p(\widehat \U_n)]
\le \Exp[L_p(\widehat \V_n)] + O \left( \Exp \left[ \sum_{\X \in \widehat \U_n} \|\X - \b_\X\|^p \right] \right) \; .
\end{align*}
To upper the right hand side we use (\ref{equation:boundary-edges}) and (\ref{equation:face-graph}),
which proves that $\Exp[L_p(\widehat \U_n)] \le O(\max(n^{1-p/d-1/d},1))$ and finishes the proof.
\end{proof}

\section{Concentration and Estimator of Entropy}
\label{section:concentration}

In this section, we show that if $\V_n$ is a set of $n$ points drawn i.i.d. from
any distribution over $[0,1]^d$ then $L_p(\V_n)$ is tightly concentrated. That
is, we show that with high probability $L_p(\V_n)$ is within
$O(n^{1/2-p/(2d)})$ its expected value. We use this result at the end of this
section to give a proof of Theorem~\ref{thm:mainentropy}.

It turns out that in order to derive the concentration result, the properties
of the distribution generating the points are irrelevant (even the existence of
density is not necessary). The
only property that we exploit is smoothness of $L_p$.  As a technical tool, we
use the isoperimetric inequality for Hamming distance and product measures.
This inequality is, in turn, a simple consequence of Talagrand's isoperimetric
inequality, see
e.g.~\cite{Dubhashi-Panconesi2009,Alon-Spencer2000,Talagrand1995}.  To phrase
the isoperimetric inequality, we use Hamming distance $H(\x_{1:n},\y_{1:n})$
between two tuples $\x_{1:n}=(\x_1, \x_2, \dots, \x_n)$, $\y_{1:n}=(\y_1, \y_2,
\dots, \y_n)$ which is defined as the number of elements in which $\x_{1:n}$ and
$\y_{1:n}$ disagree.

\begin{theorem}[Isoperimetric Inequality]
Let $A \subset \Omega^n$ be a subset of an $n$-fold product of a probability space
equipped with a product measure. For any $t \ge 0$ let
$A_t = \left\{ \x_{1:n} \in \Omega^n ~:~ \exists \y_{1:n} \in \Omega^n \ \text{s.t.} \ H(\x_{1:n},\y_{1:n}) \le t \right\}$ be an expansion of $A$.
Then, for any $t \ge 0$,
$$
\Pr[A] \Pr[\overline{A_t}] \le \exp \left( - \frac{t^2}{4n} \right) \; ,
$$
where $\overline{A_t}$ denotes the complement of $A_t$ with respect to $\Omega^n$.
\end{theorem}

\begin{theorem}[Concentration Around the Median]
\label{theorem:concentration}
Let $\V_n$ consists of $n$ points drawn i.i.d. from an absolutely continuous probability
distribution over $[0,1]^d$, let $0 \le p \le d$. For any $t > 0$,
$$
\Pr \left[ |L_p(\V_n) - M(L_p(\V_n))| > t \right] \quad \le \quad e^{-\Theta(t^{2d/(d-p)}/n)} \; ,
$$
where $M(\cdot)$ denotes the median of a random variable.
\end{theorem}

\begin{proof}
Let $\Omega=[0,1]^d$ and $\V_n = \{\X_1, \X_2, \dots, \X_n\}$, where $\X_1, \X_2, \dots, \X_n$
are independent. To emphasize that we are working in a product space,
we use the notations $L_p(\x) := L_p(\{x_1, x_2, \dots, x_n\})$,
$L_p(\X_{1:n}) := L_p(\V_n) = L_p(\{\X_1, \X_2, \dots, \X_n\})$ and $M := M(L_p(\X_{1:n}))$.
Let $A=\{ \x \in \Omega^n ~:~ L_p(\x) \le M \}$.
By smoothness of $L_p$ there exists a constant $C > 0$ such that
$$
L_p(\x) \le L_p(\y) + C \cdot H(\x,\y)^{1-p/d} \; .
$$
Therefore, $L_p(\x) > M + t$ implies that $\x \in \overline{A_{(t/C)^{d/(d-p)}}}$. Hence
for a random $\X_{1:n} = (\X_1, \X_2, \dots, \X_n)$
$$
\Pr[L_p(\X_{1:n}) > M + t]
\le \Pr[\X_{1:n} \in \overline{A_{(t/C)^{d/(d-p)}}}]
\le \frac{1}{\Pr[A]} \ e^{-\Theta(t^{2d/(d-p)}/n)}
$$
by the isoperimetric inequality.
Similarly, we set $B=\overline{A}$ and note that by smoothness we have also the reversed inequality
$$
L_p(\y) \le L_p(\x) + C \cdot H(\x,\y)^{1-p/d} \; .
$$
Therefore, $L_p(\x) < M + t$ implies that $\x \in \overline{B_{(t/C)^{d/(d-p)}}}$. By the same argument as before
$$
\Pr[L_p(\X_{1:n}) < M + t]
\le \Pr[\X_{1:n} \in \overline{B_{(t/C)^{d/(d-p)}}}]
\le \frac{1}{\Pr[B]} \ e^{-\Theta(t^{2d/(d-p)}/n)}  \; .
$$
The theorem follows by the union bound and the fact that $\Pr[A]=\Pr[B]=1/2$.
\end{proof}

\begin{corollary}[Deviation of the Mean and the Median] \label{corollary:deviation-mean-median}
Let $\V_n$ consists of $n$ points drawn i.i.d. from an absolutely continuous probability
distribution over $[0,1]^d$, let $0 \le p \le d$ and $S \subset \N^+$ a finite set. Then
$$
\left| \Exp[L_p(\V_n)] - M(L_p(\V_n)) \right| \le O(n^{1/2-p/(2d)}) \; .
$$
\end{corollary}

\begin{proof}
For conciseness let $L_p = L_p(\V_n)$ and $M = M(L_p(\V_n))$.
We have
\begin{align*}
|\Exp[L_p] - M| 
& \le \Exp|L_p-M| \\
& = \int_{0}^\infty \Pr[|L_p - M| > t] \ \ud t \\
& \le \int_{0}^\infty e^{-\Theta(t^{2d/(d-p)}/n)} \ud t \\
& = \Theta(n^{1/2-p/(2d)}) \; . \qedhere
\end{align*}
\end{proof}

Putting these pieces together we arrive at what we wanted to prove:

\begin{corollary}[Concentration]
Let $\V_n$ consists of $n$ points drawn i.i.d. from an absolutely continuous probability
distribution over $[0,1]^d$, let $0 \le p \le d$ and $S \subset \N^+$ and finite.
For any $\delta
> 0$ with probability at least $1-\delta$,
\begin{eqnarray}
\left| \Exp[L_p(\V_n)] \ - L_p(\V_n) \right| \quad \le \quad O(n\log(1/\delta))^{1/2-p/(2d)} \; . \label{equation:final_concentration}
\end{eqnarray}
\end{corollary}

\begin{proof}[Proof of Theorem~\ref{thm:mainentropy}]
By scaling and translation, we can assume that the support of $\mu$ is contained in the unit cube
$[0,1]^d$. The first part of the theorem follows
immediately from Theorem \ref{theorem:Redmond-Yukich}. To prove the second part
observe from \eqref{equation:final_concentration} that for any $\delta
> 0$ with probability at least $1-\delta$,
\begin{eqnarray}
\left| \frac{\Exp[L_p(\V_n)]}{\gamma n^{1-p/d}} \ - \frac{L_p(\V_n)}{\gamma n^{1-p/d}} \right| \quad \le \quad O\left(n^{-1/2+p/(2d)} (\log(1/\delta))^{1/2-p/(2d)}\right) \; .
\label{equation:final_concentration_normalized}
\end{eqnarray}
It is easy to see that if $0<p \leq d-1$ then $-1/2+p/(2d)<-\frac{d-p}{d(2d-p)}<0$, and if
$d-1 \leq p <d$ then $-1/2+p/(2d)<-\frac{d-p}{d(d+1)}<0$.
Now using \eqref{equation:final_concentration_normalized}, Theorem~\ref{theorem:Koo-Lee} and the triangle inequality, we have that for any $\delta
> 0$ with probability at least $1-\delta$,
\begin{align*}
\left| \frac{L_p(\V_n)}{\gamma n^{1-p/d}} \ - \int_{[0,1]^d} f^{1-p/d}(\x) \ \ud \x \right| 
& \le \left| \frac{\Exp[L_p(\V_n)]}{\gamma n^{1-p/d}} \ - \frac{L_p(\V_n)}{\gamma n^{1-p/d}} \right| \\
& \qquad + \left| \frac{\Exp[L_p(\V_n)]}{\gamma n^{1-p/d}} \ -\int_{[0,1]^d} f^{1-p/d}(\x) \ \ud \x \right| \nonumber \\
&\le \begin{cases}
O\left( n^{-\frac{d-p}{d(2d-p)}}(\log(1/\delta))^{1/2-p/(2d)} \right), & \text{if $0 < p < d-1$}\;; \\
O\left( n^{-\frac{d-p}{d(d+1)}}(\log(1/\delta))^{1/2-p/(2d)} \right), & \text{if $d-1 \le p < d$} \; .
\end{cases}
\end{align*}
To finish the proof of \eqref{equation:entropy-rate} exploit the fact that
$\log(1\pm x) = \pm O(x)$ for $x \to 0$.
%
%
\end{proof}

\section{Copulas and Estimator of Mutual Information}
\label{section:copula-MI-appendix}

The goal of this section is to prove Theorem~\ref{thm:maininformation} on
convergence of the estimator $\widehat I_{\alpha}$. The main additional problem
that we need to deal with in the proof is the effect of the empirical copula
transformation.  A version of the classical Kiefer-Dvoretzky-Wolfowitz theorem
due to Massart gives a convenient way to do it; see
e.g.~\cite{Devroye-Lugosi2001}.

\begin{theorem}[Kiefer-Dvoretzky-Wolfowitz]
Let $X_1, X_2, \dots, X_n$ be an i.i.d. sample from a probability distribution over $\R$
with c.d.f. $F:\R \to [0,1]$.
Define the \emph{empirical c.d.f.}
$$
\widehat F(x) = \frac{1}{n} | \{ i ~:~ 1 \le i \le n, \ X_i \le x \} | \qquad \text{for } x \in \R \; .
$$
Then, for any $t \ge 0$,
$$
\Pr \left[ \sup_{x \in \R} |F(x) - \widehat F(x)| > t \right] \le 2 e^{-2nt^2} \; .
$$
\end{theorem}

As a simple consequence of the Kiefer-Dvoretzky-Wolfowitz theorem, we can derive that $\widehat \F$
is a good approximation of $\F$.
\begin{lemma}[Convergence of Empirical Copula]
Let $\X_1, \X_2, \dots, \X_n$ be an i.i.d. sample from a probability distribution
over $\R^d$ with marginal c.d.f.'s $F_1, F_2, \dots, F_d$. Let $\F$ be the copula
defined by \eqref{equation:copula-transformation} and let $\widehat \F$
be the empirical copula transformation defined by~\eqref{equation:empirical-copula-transformation}.
Then, for any $t \ge 0$,
$$
\Pr \left[ \sup_{\x \in \R^d} \|\F(\x) - \widehat \F(\x)\|_2 > t \right] \le 2d e^{-2ndt^2} \; .
$$
\end{lemma}

\begin{proof}
Using $\|\cdot\|_2 \le \sqrt{d}\|\cdot\|_{\infty}$ in $\R^d$ and union-bound we have
\begin{align*}
\Pr \left[ \sup_{\x \in \R^d} \|\F(\x) - \widehat \F(\x)\|_2 > t \right]
& \le \Pr \left[ \sup_{\x \in \R^d} \|\F(\x) - \widehat \F(\x)\|_\infty > t \sqrt{d} \right] \\
& = \Pr \left[ \sup_{x \in \R} \max_{1 \le j \le d} |F_j(x) - \widehat F_j(x)| > t \sqrt{d} \right] \\
& \le \sum_{i=1}^d \Pr \left[ \sup_{x \in \R} |F_j(x) - \widehat F_j(x)| > t \sqrt{d} \right] \\
& \le 2de^{-2ndt^2} \; . \qedhere
\end{align*}
\end{proof}

The following corollary is an obvious consequence of this lemma:
\begin{corollary}[Convergence of Empirical Copula]
\label{corollary:Empirical_copula_deviation}
Let $\X_1, \X_2, \dots, \X_n$ be an i.i.d. sample from a probability distribution
over $\R^d$ with marginal c.d.f.'s $F_1, F_2, \dots, F_d$. Let $\F$ be the copula
defined by \eqref{equation:copula-transformation}, and let $\widehat \F$
be the empirical copula transformation defined by~\eqref{equation:empirical-copula-transformation}. Then, for any $\delta>0$,
\begin{eqnarray}
\Pr\left[ \max_{1 \le i \le n} \|\F(\X_i) - \widehat \F(\X_i) \| < \sqrt{\frac{\log(2d/\delta)}{2nd}} \right]
\ge 1-\delta \; .
\end{eqnarray}
\end{corollary}

\begin{proposition}[Order statistics]
\label{proposition:order-statistics}
Let $a_1,a_2,\ldots,a_m$ and $b_1,b_2,\ldots,b_m$ be real numbers.
Let $a_{(1)}\le a_{(2)}\le\ldots\le a_{(m)}$ and $b_{(1)}\le b_{(2)}\le\ldots \le b_{(m)}$ be the same numbers sorted in ascending order. Then,  $|a_{(i)}-b_{(i)}|\le \max_j|a_j-b_j|$, for all $1 \le i \le m$.
\end{proposition}

\begin{proof}
The proof is left as an exercise for the reader.
\end{proof}

\begin{lemma}[Perturbation]\label{lemma:Perturbation}
Consider points $\x_1, \x_2, \dots, \x_n, \y_1, \y_2, \dots, \y_n \in \R^d$ such that
$\|\x_i - \y_i\| < \epsilon$ for all $1 \le i \le n$. Then,
$$
|L_p(\{\x_1, \x_2, \dots, \x_n\}) - L_p(\{\y_1, \y_2, \dots, \y_n\})| \le
\begin{cases}
O\left( n\epsilon^p\right), & \text{if $0 < p < 1$}\;; \\
O\left( n\epsilon\right), & \text{if $1 \le p $} \; .
\end{cases}
$$
\end{lemma}

\begin{proof}
Let $k=\max S$, $A = \{\x_1, \x_2, \dots, \x_n\}$ and $B = \{\y_1, \y_2, \dots, \y_n\}$.
Let $w_A(i,j) = \|\x_i - \x_j\|^p$ and $w_B(i,j) = \|\y_i - \y_j\|^p$ be the edge weights defined by $A$ and  $B$
respectively.
Let $a^i_{(j)}$ be the $p$-th power of the distance from $\x_i$ to its $j$-th nearest-neighbor in $A$, for $1\le i \le n$ ,$1\le j\le n-1$. Similarly, let $b^i_{(j)}$ be the $p$-th power of the distance from $\y_i$ to its $j$-th nearest-neighbor in $B$. Note that for any $i$, if we sort the real numbers $w_A(i,1),\ldots,w_A(i,i-1),w_A(i,i+1),\ldots,w_A(i,n)$, then we get
$a^i_{(1)}\le a^i_{(2)}\le \ldots\le a^i_{(n-1)}$. Similarly for $w_B$'s and $b^i_{(j)}$'s.
Using these notations we can write
\begin{align*}
|L_p(A) - L_p(B)|
& = \left|\sum_{i=1}^n \sum_{j \in S} a^i_{(j)}-b^i_{(j)}\right| \\
& \le \sum_{i=1}^n \sum_{j \in S} \left|a^i_{(j)}-b^i_{(j)}\right| \\
& \le \sum_{i=1}^n \sum_{j \in S} \max_{1 \le i,j \le n}\left|a^i_{(j)}-b^i_{(j)}\right| \\
& \le \sum_{i=1}^n \sum_{j \in S}\max_{i,j}\left|w_A(i,j)-w_B(i,j)\right| \\
& \le kn \max_{1 \le i,j \le n} \left| w_A(i,j) - w_B(i,j) \right| \; .
\end{align*}
The third inequality follows from Proposition~\ref{proposition:order-statistics}.
It remains to bound $|w_A(i,j) - w_B(i,j)|$. We consider two cases:

\emph{Case $0 < p < 1$.}
Using $|u^p-v^p| \le |u-v|^p$ valid for any $u,v \ge 0$ and
the triangle inequality
\begin{equation}
\label{equation:doubled-sided-triangle}
\bigg| \|\textbf{a} - \textbf{b}\| - \|\textbf{c} - \textbf{d}\| \bigg|
\le \|\textbf{a} - \textbf{c}\| + \|\textbf{b} - \textbf{d}\|
\end{equation}
valid for any $\textbf{a},\textbf{b},\textbf{c},\textbf{d} \in \R^d$ we have
\begin{align*}
|w_A(i,j) - w_B(i,j)|
& =  \left| \|\x_i - \x_j\|^p - \|\y_i - \y_j\|^p \right| \\
& \le \left| \|\x_i - \x_j\| - \|\y_i - \y_j\| \right|^p \\
& \le \left(\|\x_i - \y_i\| + \|\x_j - \y_j\|\right)^p \\
& \le 2^p \epsilon^p \; .
\end{align*}

\emph{Case $p \ge 1$.}
Consider the function $f(u) = u^p$ on interval $[0,\sqrt{d}]$. On this interval $|f'(u)| \le pd^{(p-1)/2}$
and so $f$ is Lipschitz with constant $pd^{(p-1)/2}$. In other words, for any $u,v \in [0,\sqrt{d}]$,
$|u^p - v^p| \le pd^{(p-1)/2}|u-v|$. Thus
\begin{align*}
|w_A(i,j) - w_B(i,j)|
& =  \left| \|\x_i - \x_j\|^p - \|\y_i - \y_j\|^p \right| \\
& \le pd^{(p-1)/2} \left| \|\x_i - \x_j\| - \|\y_i - \y_j\| \right| \\
& \le pd^{(p-1)/2} (\|\x_i - \y_i\| + \|\x_j - \y_j\|) \\
& \le 2\epsilon pd^{(p-1)/2} \; ,
\end{align*}
where the second inequality follows from~(\ref{equation:doubled-sided-triangle}).
\end{proof}

\begin{corollary}[Copula Perturbation] \label{corollary:copula-perturbation}
Let $\X_1, \X_2, \dots, \X_n$ be an i.i.d. sample from a probability distribution
over $\R^d$ with marginal c.d.f.'s $F_1, F_2, \dots, F_d$. Let $\F$ be the copula
defined by \eqref{equation:copula-transformation} and let $\widehat \F$
be the empirical copula transformation defined by~(\ref{equation:empirical-copula-transformation}). Let $\Z_i=\F(\X_i)$ and
$\widehat{\Z}_i=\widehat \F(\X_i)$.
Then for any $\delta>0$,
with probability at least $1-\delta$,
$$
\left|\frac{L_p(\Z_{1:n})}{\gamma n^{1-p/d}} - \frac{L_p(\widehat{\Z}_{1:n})}{\gamma n^{1-p/d}}\right| \le
\begin{cases}
O\left( n^{p/d-p/2}(\log(1/\delta))^{p/2}\right), & \text{if $0 < p < 1$}\;; \\
O\left( n^{p/d-1/2}(\log(1/\delta))^{1/2}\right), & \text{if $1 \le p $} \; .
\end{cases}
$$
\end{corollary}

\begin{proof}
It follows immediately from Corollary~\ref{corollary:Empirical_copula_deviation} and Lemma~\ref{lemma:Perturbation} that with probability at least $1-\delta$,
$$
|L_p(\{\Z_{1:n}\}) - L_p(\widehat{\Z}_{1:n})| \le
\begin{cases}
O\left( n^{1-p/2}(\log(1/\delta))^{p/2}\right), & \text{if $0 < p < 1$}\;; \\
O\left( n^{1/2}(\log(1/\delta))^{1/2}\right), & \text{if $1 \le p $} \; .
\end{cases}
$$
\end{proof}

We are now ready to give the proof of Theorem~\ref{thm:maininformation}.

\begin{proof}[Proof of Theorem~\ref{thm:maininformation}]
Let $g$ denote the density of the copula of $\mu$.
The first part follows from \eqref{equation:entropy-as}, Corollary~\ref{corollary:copula-perturbation} and a standard Borel-Cantelli argument with $\delta=1/n^2$. Corollary~\ref{corollary:copula-perturbation} puts the restrictions $d\ge 3$ and $1/2<\alpha<1$.

The second part can be proved along the same lines. From \eqref{equation:entropy-rate} we have that
for any $\delta>0$ with probability at least $1-\delta$,
\begin{eqnarray*}
\left| \frac{L_p(\Z_{1:n})}{\gamma n^{1-p/d}} \ - \int_{[0,1]^d} g^{1-p/d}(\x) \ \ud \x \right|
&\le&\begin{cases}
O\left( n^{-\frac{d-p}{d(2d-p)}}(\log(1/\delta))^{1/2-p/(2d)} \right), & \text{if $0 < p < d-1$} \;;\\
O\left( n^{-\frac{d-p}{d(d+1)}}(\log(1/\delta))^{1/2-p/(2d)} \right), & \text{if $d-1 \le p < d$} \; .
\end{cases}
\end{eqnarray*}
Hence using the triangle inequality again, and exploiting that $(\log(1/\delta))^{1/2-p/(2d)}<(\log(1/\delta))^{1/2}$ if $0 <p$, $\delta < 1$,
 we have that with probability at least $1-\delta$,
$$
\left|  \frac{L_p(\widehat{\Z}_{1:n})}{\gamma n^{1-p/d}} \ - \int_{[0,1]^d} g^{1-p/d}(\x) \ \ud \x \right|
\le\begin{cases}
O\left( \max\{n^{-\frac{d-p}{d(2d-p)}},n^{-p/2+p/d} \}
\sqrt{\log(1/\delta)} \right), & \text{if $0 < p \le 1$}\;; \\
O\left( \max\{n^{-\frac{d-p}{d(2d-p)}},n^{-1/2+p/d}\}\sqrt{\log(1/\delta)} \right),  & \text{if $1 \le p \le d-1$}\;; \\
O\left( \max\{n^{-\frac{d-p}{d(d+1)}},n^{-1/2+p/d}\}\sqrt{\log(1/\delta)} \right),  & \text{if $d-1 \le p <d$}\;.
\end{cases}
$$
To finish the proof exploit that when $x\to 0$ then $\log(1 \pm x)=\pm O(x)$.
\end{proof}

\end{document}